\documentclass[10pt,a4paper,twoside]{article}

\usepackage[utf8]{inputenc}
\usepackage[table,xcdraw]{xcolor}
\usepackage{empheq}

\usepackage{geometry}
\usepackage{float}
\usepackage{subfigure}
\usepackage{subfiles}
\usepackage{xr}

\usepackage{amsmath,amssymb,amsfonts}
\usepackage{mathtools}
\usepackage{mathrsfs}
\usepackage{makeidx}
\usepackage{idxlayout} 
\usepackage{multicol} 
\usepackage{multirow}

\usepackage{exscale,relsize} 
\newcommand{\mypound}{\scalebox{0.8}{\raisebox{0.4ex}{\#}}}

\usepackage{mleftright}
\usepackage{theorem}
\usepackage{graphics}                 
\usepackage{accents}
\usepackage{kbordermatrix}
\usepackage[margin=10pt,font=small,labelfont=bf]{caption}

\usepackage{mdframed}
\usepackage{verbatim}

\usepackage{cases}
\usepackage{makeidx}
\usepackage{titlesec}
\usepackage{tabu}

\usepackage{algorithmic}
\usepackage{algorithm}

\usepackage{tikz}
\usetikzlibrary{arrows,shapes}
\usepackage{url}

\usepackage{bm,xstring}
\usepackage{xfakebold}

\newcommand{\fbseries}{\unskip\setBold\aftergroup\unsetBold\aftergroup\ignorespaces}
\makeatletter
\newcommand{\setBoldness}[1]{\def\fake@bold{#1}}
\makeatother

\def\Pext{\mathbf{P}_{\!\!\mathrm{ext}}}

\newcommand{\mat}[1]{\mathbf{#1}}

\newcommand{\argmin}{\mathop{\mathrm{arg\,min}}}

\newcommand{\minimize}{\mathop{\mathrm{minimize}}}

\newcommand{\subjectto}{\mathop{\mathrm{subject\,to}}}

\newcommand{\myDelta}{{\textstyle \mathsmaller{\varDelta}}} 

\newcommand{\dsum}{\displaystyle\sum}

\newcommand{\cditto}{\raisebox{.5ex}{\hbox to 2em{\hrulefill}}}

\mleftright 
\title{Relative Entropy-Regularized Optimal Transport\\
   on a Graph: a new algorithm and an experimental comparison}

\author{Sylvain Courtain, Guillaume Guex, Ilkka Kivimaki \& Marco Saerens}

\begin{document}

\setlength{\parskip}{1pt plus 1pt minus 1pt} 

\sloppy 

\date{}
\maketitle

\begin{abstract}
Following \cite{Guex-2016,Guex-2019}, the present work investigates a new relative entropy-regularized algorithm for solving the optimal transport on a graph problem\footnote{Called the minimum cost flow problem in operation research \cite{Ahuja-1993}.} within the randomized shortest paths formalism. More precisely, a unit flow is injected into a set of input nodes and collected from a set of output nodes while minimizing the expected transportation cost together with a paths relative entropy regularization term, providing a randomized routing policy. The main advantage of this new formulation is the fact that it can easily accommodate edge flow capacity constraints which commonly occur in real-world problems. The resulting optimal routing policy, i.e., the probability distribution of following an edge in each node, is Markovian and is computed by constraining the input and output flows to the prescribed marginal probabilities thanks to a variant of the algorithm developed in \cite{Courtain-2020}. In addition, experimental comparisons with other recently developed techniques show that the distance measure between nodes derived from the introduced model provides competitive results on semi-supervised classification tasks.
\end{abstract}

\section{Introduction}
\label{Sec_introduction01}

\subsection{General introduction}

This work studies a flow-constrained randomized shortest paths \cite{Kivimaki-2012,Saerens-2008,Yen-08K} formulation to the relative entropy-regularized, or randomized, optimal transport problem on a graph with multiple inputs and outputs having fixed marginal probabilities (or margins, providing fixed unit input and output flows), studied in \cite{Guex-2016,Guex-2019}. This last work extended the relative entropy-regularized optimal transport problem  (see the recent work \cite{Cuturi2013}, but also \cite{Erlander-1990,Kapur-1992,Wilson-1970}) to a graph structure. Intuitively, it aims to carry goods from some input nodes to output nodes with least expected cost while maintaining a prescribed level of relative entropy of the path probabilities connecting inputs to outputs. In this problem, the input flows (proportion of goods carried from input nodes) and output flows (proportion of goods carried from output nodes) are constrained to be equal to some predefined values (marginal probabilities), which are not defined in the standard \textbf{randomized shortest paths} (RSP, \cite{Kivimaki-2012,Saerens-2008,Yen-08K}) and \textbf{bag-of-paths} (BoP, \cite{Francoisse-2017,Mantrach-2009}) models. The introduced model will therefore be called the \textbf{margin-constrained bag-of-paths} model along the paper, in order to remain consistent with \cite{Guex-2019}.

The introduced algorithm solving this problem provides an optimal randomized policy balancing exploitation and exploration through a simple iterative algorithm inspired by \cite{Courtain-2020}. Similarly to the standard randomized shortest paths and bag-of-paths frameworks\footnote{The main difference between the BoP and the RSP formalism is that, for the BoP, all possible paths in the network are considered \cite{Francoisse-2017,Mantrach-2009}, whereas only source-target paths connecting two nodes of interest are considered in the RSP \cite{Kivimaki-2012,Saerens-2008,Yen-08K}. The RSP therefore avoids the need for defining prior distributions on source and on target nodes because there is only one single source and target.}, the model is monitored by a parameter $\theta$ in such a way that, when $\theta$ goes to infinity, it approximates the optimal, lowest-cost, solution to the transportation problem. Conversely, when $\theta$ is close to zero, the solution approaches a random walk behavior (provided a priori by the user) in terms of relative entropy (also called Kullback-Leibler divergence). Thus, when varying $\theta$, the model interpolates between an optimal (exploitation) and a random (exploration) behavior.

The main idea, in comparison with the previous work (mainly \cite{Guex-2016,Guex-2019}), is the following. The relative entropy-regularized optimal transport on a graph problem studied in \cite{Guex-2019} is based on a BoP formulation where the set of all paths between the source and target nodes is considered. On the contrary, in the present work, the problem is rephrased within a RSP formalism (\cite{Saerens-2008}; inspired by \cite{Akamatsu-1996}) only considering the set of paths from one single source supernode and one target supernode, both added to the original network.

Furthermore, this rephrasing into a source-target RSP problem allows us to easily define capacity constraints on edge flows, as presented in \cite{Courtain-2020}. Flow capacity constraints are relevant in many graph-based applications; for instance, in the case of traffic or passenger volume constraints on transportation networks \cite{Ahuja-1993}. This previous work \cite{Courtain-2020} allows us to reformulate the margin-constrained BoP on a graph problem into a capacity constrained RSP problem which has its own merits, for instance allowing additional flow capacity constraints. As an application, the margin-constrained bag-of-paths surprisal distance measure between nodes \cite{Guex-2019} is derived within the new formalism.

To summarize, because the problem is transformed into a RSP single source-single target framework, it suffices to follow the procedure introduced in \cite{Courtain-2020} for dealing with both the margin and the capacity constraints. The new algorithm therefore solves the relative entropy-regularized minimum expected cost flow with capacity constraints problem on a graph. It is, however, slower than the dedicated algorithm developed in \cite{Guex-2019} so that it should only be used when capacity constraints are indeed present.

In addition to the introduction of this new algorithm, we also conducted an extensive experimental comparison between the newly introduced model and other state-of-the-art models on graph-based semi-supervised classification problems. This experimental comparison was left for further work in \cite{Guex-2019}, where a structural distance measure between pairs of nodes was defined. The results of this comparison suggest that the distance measure derived from the margin-constrained BoP model is clearly competitive with respect to the other investigated distance measures.

In order to avoid redundancy, the present paper will focus on the derivation of the new model and the experimental comparison. For a comprehensive discussion and related work concerning the margin-constrained BoP problem on a graph, as well as some illustrative examples, see \cite{Guex-2016,Guex-2019}.

\subsection{Main contributions and content}

The main contributions of this work are
\begin{itemize}
    \item The introduction of a new algorithm for solving the relative entropy-regularized optimal transport (in terms of minimum expected cost flow) on a graph problem by considering edge flow constraints.
    \item An experimental comparison between the distance measure derived from the studied model and other state-of-the-art distance measures between nodes on a graph on semi-supervised classification problems.
\end{itemize}

The next section presents the notation and some preliminaries. Section \ref{Sec_optimal_transport01} develops the new RSP-based algorithm relying on flow constraints. Then, Section \ref{Sec_Experimental_Comparison01} details the experimental comparison and its results. Section \ref{Sec_Conclusion01} presents the conclusions of the work.

\section{Notation, problem statement, and preliminaries}
\label{Sec_notation_problem_statement01}

\subsection{Notation}
\label{Subsec_notation01}

Assume a directed, strongly connected and weighted, graph $G = \{\mathcal{V},\mathcal{E}\}$ containing $(n-2)$ nodes\footnote{See later for the justification; an extended graph with two additional nodes, and thus $n$ nodes in total, will be defined later in Subsection \ref{Subsec_extended_graph01}.}, in which we have to carry goods from a predefined set of input nodes $\mathcal{I}n$ to a set of output nodes $\mathcal{O}ut$. The user specifies the proportions of non-negative, continuous, \textbf{flow} $\sigma_{i}^{\mathrm{in}}$ coming from each input $i \in \mathcal{I}n$ as well as the proportion of flow delivered to each output, $\sigma_{j}^{\mathrm{out}}$, $j \in \mathcal{O}ut$. For all other nodes $i \notin \mathcal{I}n$, we set $\sigma_{i}^{\mathrm{in}} = 0$. Symmetrically, for all nodes $j \notin \mathcal{O}ut$, $\sigma_{j}^{\mathrm{out}} = 0$. In general, we almost always have that input nodes are different from output nodes, $\mathcal{I}n \cap \mathcal{O}ut = \varnothing$, but this assumption is not needed in the model.

Let us further assume
\begin{equation}
\begin{dcases}
\sum_{i \in \mathcal{I}n} \sigma_{i}^{\mathrm{in}}   = 1 &\text{with all } \sigma_{i}^{\mathrm{in}} \ge 0 \\
\sum_{j \in \mathcal{O}ut} \sigma_{j}^{\mathrm{out}} = 1 &\text{with all } \sigma_{j}^{\mathrm{out}} \ge 0
\end{dcases}
\label{Eq_flow_constraints01}
\end{equation}
In other words, a unit flow is considered.
If we have to transport non-unitary flow, we simply find the solution for a unit flow and then multiply the quantities by the total flow. The $n \times 1$ column vector containing the input flow of each node is denoted by $\boldsymbol{\sigma}_{\mathrm{in}}$ whereas the vector containing the output flows is $\boldsymbol{\sigma}_{\mathrm{out}}$.

Moreover, it is assumed that a non-negative \textbf{local cost} $c_{ij}$ is associated to each edge $(i,j)$, reflecting the penalty of following this edge (it can be a distance, a cost, a travel time, etc). The \textbf{cost matrix} $\mathbf{C}$ contains the individual costs $c_{ij}$ as elements. When there is no edge from node $i$ to node $j$ we consider the cost $c_{ij}$ to be infinite. The graph $G$ is also associated to an \textbf{adjacency matrix} $\mathbf{A}$ containing local affinities, or weights, $\{ a_{ij} \}$ between nodes. When there is no direct link between two nodes $i,j$, $a_{ij} = 0$. Matrices $\mathbf{C}$ and $\mathbf{A}$ are given and depend on the problem at hand. The \textbf{transition probability matrix} associated to the natural random walk on this graph $G$ is $\mathbf{P}$, with elements
\begin{equation}
[\mathbf{P}]_{ij} = p_{ij} = \dfrac{a_{ij}}{\dsum_{j'\in \mathcal{S}ucc(i)} a_{ij'}}
= \dfrac{a_{ij}}{\dsum_{j'=1}^{n} a_{ij'}}
\label{Eq_transition_probabilities_original_graph01}
\end{equation}
where $\mathcal{S}ucc(i)$ is the set of successor nodes of node $i$. The third equality is valid because the elements on the $i$th row of the adjacency matrix are equal to $0$ for the missing links $j \in \mathcal{E} \setminus \mathcal{S}ucc(i)$. Finally, we assume that the Markov chain associated to the random walk on $G$ is \textbf{regular}, that is, strongly connected and aperiodic\footnote{This property is needed in Appendix \ref{Subsec_appendix_A2} for computing the transition matrix of the natural random walk on the extended graph, but this assumption could probably be alleviated.}.

\subsection{Problem statement}
\label{Subsec_problem_statement01}

The problem is then to find the best policy, which takes the form of path probabilities, for carrying the input flow from the source nodes to the target nodes, minimizing the expected cost along the paths connecting sources to targets, while 
\begin{enumerate}
\item keeping a given level of exploration quantified by the Kullback-Leibler divergence between path probabilities and complete random paths provided by the natural random walk (\ref{Eq_transition_probabilities_original_graph01}), and
\item satisfying the flow constraints stating that input flows and output flows are fixed to $\sigma_{i}^{\mathrm{in}}$ for each node $i \in \mathcal{I}n$ and $\sigma_{j}^{\mathrm{out}}$ for each node $j \in \mathcal{O}ut$.
\end{enumerate}
\noindent
This problem will be solved in Section \ref{Sec_optimal_transport01}; before, let us introduce an extended graph $G_{\mathrm{ext}}$ by adding two nodes to $G$.

\definecolor{myRed}{RGB}{139,0,0}
\definecolor{myGreen}{RGB}{0,100,0}
\begin{figure}[t]
    \centering
    \subfigure{\begin{tikzpicture}[shorten >=1pt,auto, ultra thick,
  node_style/.style={circle,ball color = white,font=\sffamily\bfseries,minimum size=0.8cm,draw=black},
  node_style_in/.style={circle,ball color = green,font=\sffamily\bfseries,minimum size=0.8cm,draw=black},
  node_style_out/.style={circle,ball color = red,font=\sffamily\bfseries,minimum size=0.8cm,draw=black},
  scale=0.85,transform shape]

    \node[node_style_in]  (1) at (-2,1) {2};
    \node[node_style_in]  (2) at (-2,-1) {3};
    
    \node[node_style]  (3) at (0,2) {4};
    \node[node_style]  (4) at (0,0) {5};
    \node[node_style]  (5) at (0,-2) {6};
    
    \node[node_style_out]  (6) at (2,1) {7};
    \node[node_style_out]  (7) at (2,-1) {8};

    \draw[draw=black,->,>=stealth']   (1) edge node{} (2);
    \draw[draw=black,->,>=stealth']   (1) edge node{} (3);
    \draw[draw=black,->,>=stealth']   (1) edge node{} (4);
    \draw[draw=black,->,>=stealth']   (2) edge node{} (4);
    \draw[draw=black,->,>=stealth']   (2) edge node{} (5);
    \draw[draw=black,->,>=stealth']   (4) edge node{} (3);
    \draw[draw=black,->,>=stealth']   (3) edge node{} (6);
    \draw[draw=black,->,>=stealth']   (4) edge node{} (5);
    \draw[draw=black,->,>=stealth']   (4) edge node{} (6);
    \draw[draw=black,->,>=stealth']   (4) edge node{} (7);
    \draw[draw=black,->,>=stealth']   (5) edge node{} (7);
    \draw[draw=black,->,>=stealth']   (7) edge node{} (6);

\end{tikzpicture}}
    \subfigure{\begin{tikzpicture}[shorten >=1pt,auto, ultra thick,
  node_style/.style={circle,ball color = white,font=\sffamily\bfseries,minimum size=0.8cm,draw=black},
  node_style_in/.style={circle,ball color = green,font=\sffamily\bfseries,minimum size=0.8cm,draw=black},
  node_style_out/.style={circle,ball color = red,font=\sffamily\bfseries,minimum size=0.8cm,draw=black},
  node_style_superout/.style={regular polygon,ball color = myRed,font=\sffamily\bfseries,minimum size=0.8cm,draw=black},
  node_style_superin/.style={regular polygon,ball color = myGreen,font=\sffamily\bfseries,minimum size=0.8cm,draw=black},scale=0.85,transform shape]

    \node[node_style_superin]  (0) at (-4,0) {1};
    \node[node_style_in]  (1) at (-2,1) {2};
    \node[node_style_in]  (2) at (-2,-1) {3};
    
    \node[node_style]  (3) at (0,2) {4};
    \node[node_style]  (4) at (0,0) {5};
    \node[node_style]  (5) at (0,-2) {6};
    
    \node[node_style_out]  (6) at (2,1) {7};
    \node[node_style_out]  (7) at (2,-1) {8};
    
    \node[node_style_superout]  (8) at (4,0) {9};

    \draw[draw=black,->,>=stealth']   (0) edge node{} (1);
    \draw[draw=black,->,>=stealth']   (0) edge node{} (2);
    \draw[draw=black,->,>=stealth']   (1) edge node{} (2);
    \draw[draw=black,->,>=stealth']   (1) edge node{} (3);
    \draw[draw=black,->,>=stealth']   (1) edge node{} (4);
    \draw[draw=black,->,>=stealth']   (2) edge node{} (4);
    \draw[draw=black,->,>=stealth']   (2) edge node{} (5);
    \draw[draw=black,->,>=stealth']   (4) edge node{} (3);
    \draw[draw=black,->,>=stealth']   (3) edge node{} (6);
    \draw[draw=black,->,>=stealth']   (4) edge node{} (5);
    \draw[draw=black,->,>=stealth']   (4) edge node{} (6);
    \draw[draw=black,->,>=stealth']   (4) edge node{} (7);
    \draw[draw=black,->,>=stealth']   (5) edge node{} (7);
    \draw[draw=black,->,>=stealth']   (7) edge node{} (6);
    \draw[draw=black,->,>=stealth']   (6) edge node{} (8);
    \draw[draw=black,->,>=stealth']   (7) edge node{} (8);

\end{tikzpicture}}
    \caption{On the left, a small directed graph $G$ with two input nodes $\mathcal{I}n = \{ 2,3 \}$ (in green) and two output nodes $\mathcal{O}ut = \{ 7,8\}$ (in red). On the right, the extended graph $G_{\mathrm{ext}} $ of this small directed graph $G$ with one source supernode 1 (in dark green) connected to all the input nodes and one target supernode $n=9$ (in dark red) connected to all the output nodes. Therefore, on this extended graph, the source supernode is node $1$ and the target supernode is node $n$ (the total number of nodes in  $G_{\mathrm{ext}} $).}
    \label{fig:ToyExampleGraphExtended}
\end{figure}
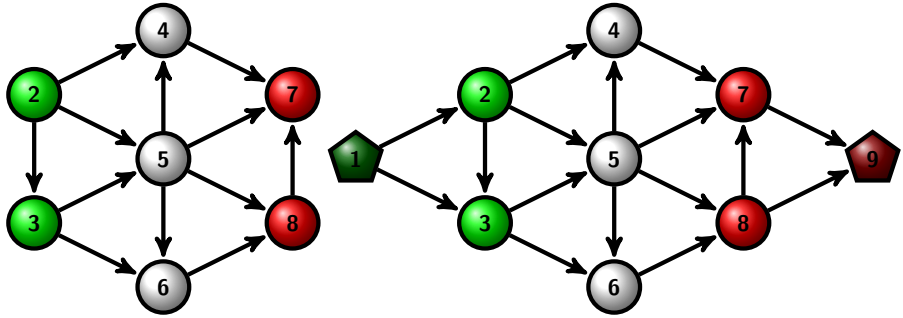

\subsection{Definition of an extended, single-source and target, graph}
\label{Subsec_extended_graph01}

For convenience, we now transform the original graph $G$ into a new, equivalent, single-source single-target graph $G_{\mathrm{ext}} = \{\mathcal{V}_{\mathrm{ext}},\mathcal{E}_{\mathrm{ext}}\}$ with $n$ nodes in a standard way \cite{Ahuja-1993,Gondran-1984}. Two nodes -- one \textbf{source node} (a supernode indexed as node $1$) and one \textbf{target node} (a supernode indexed as node $n$, where $n$ is the total number of nodes in this new, extended, graph) are added to the graph. All the other nodes in $\mathcal{V}$ (nodes $2$ to $(n-1)$) and their connections remain the same as in $G$ ($G$ is a subgraph of $G_{\mathrm{ext}}$); therefore, $\mathcal{V}_{\mathrm{ext}} = \mathcal{V} \cup \{ 1, n \}$. All nodes with label $i \in \mathcal{V}$ keep the same index numbering in $\mathcal{V}_{\mathrm{ext}}$. Moreover, the new source node (node $1$) is only connected by a directed link to the input nodes with a zero cost, and the output nodes are only connected to the new target node (node $n$) by a directed link, also with zero cost (no penalty when following these links). In order to be equivalent to the original graph $G$, the source node generates a unit flow while the target node is made killing and absorbing (a cemetery or sink node), and thus absorbs this unit flow. This new graph will be called the \textbf{extended graph}. A toy example of this concept is presented in Figure \ref{fig:ToyExampleGraphExtended}.

\subsubsection{The adjacency matrix of the extended graph}

We might now ask ourselves which weights should be associated with the edges incident to node $1$ and node $n$. A natural requirement would be that the weights of the edges incident to node $1$ are proportional to the input flows, in such a way that the corresponding transition probabilities are exactly equal to these input flows $\boldsymbol{\sigma}_{\mathrm{in}}$, as required.
However, for the nodes incident to node $n$, this is slightly more difficult. Let us denote the weights (affinities) of the edges incident to node $n$ as $\mathbf{w}$ -- these weights can either be provided explicitly by the user or can be calculated in order to have $\boldsymbol{\sigma}_{\mathrm{out}}$ as output distribution, which is developed below.

The adjacency matrix of the extended graph, $\mathbf{A}_{\mathrm{ext}}$, can be written as
\begin{equation}
 \mathbf{A}_{\mathrm{ext}} = \kbordermatrix{
               &          1 & \{ 2, \dots, (n-1) \} = \mathcal{V}                                       &         n  \cr
1              &          0 & \boldsymbol{\sigma}_{\mathrm{in}}^{\text{T}} &         0  \cr
\{ 2, \dots, (n-1) \} = \mathcal{V} & \mathbf{0} & \mathbf{A}                                           & \mathbf{w} \cr
n              &          0 & \phantom{{}^{\text{T}}} \mathbf{0}^{\text{T}}        &         0  \cr
} \label{Eq_extendedAdjacencyMatrix01}
\end{equation}

where $\mathbf{0}$ is the null column vector full of 0s. Similarly, the cost matrix becomes
\begin{equation}
 \mathbf{C}_{\mathrm{ext}} = \kbordermatrix{
               &          1 & \{ 2, \dots, (n-1) \} = \mathcal{V}                                      &         n  \cr
1              &          0 & \phantom{{}^{\text{T}}} \mathbf{0}^{\text{T}} &         0  \cr
\{ 2, \dots, (n-1) \} = \mathcal{V} & \mathbf{0} & \mathbf{C}                                           & \mathbf{0} \cr
n              &          0 & \phantom{{}^{\text{T}}} \mathbf{0}^{\text{T}}        &         0  \cr
} \label{Eq_extendedCostMatrix01}
\end{equation}

If the weights $\mathbf{w}$ are set by the user, the transition matrix $\Pext = (p_{ij}^{\mathrm{ext}})$ of the natural random walk on this extended graph $G_{\mathrm{ext}}$ can easily be computed by Equation (\ref{Eq_transition_probabilities_original_graph01}) from $\mat{A}_{\mathrm{ext}}$ instead of $\mat{A}$. In that case, our model will compute a policy (transition probabilities followed for carrying the goods) interpolating between the optimal expected lowest-cost policy and the one closest (in terms of Kullback-Leibler divergence) to the natural random walk. But it will in general, when $\theta \to 0$, \emph{not be exactly equal} to this natural random walk transition matrix because of the flow constraints in Equation (\ref{Eq_flow_constraints01}) which are \emph{not satisfied in general} for edges entering node $n$ in the extended graph $G_{\mathrm{ext}}$. 

It is, however, possible to find sets of values of the weights $\mat{w}$ such that the net flow in each of these edges is exactly equal to $\sigma_{j}^{\mathrm{out}}$ when considering a natural random walk on the extended graph \cite{Guex-2016}. Indeed, in Appendix \ref{Sec_computing_transition_matrix_consistent01}, the transition matrix on the extended graph leading to flows satisfying exactly the constraints in Equation (\ref{Eq_flow_constraints01}) for $j \in \mathcal{O}ut$ is computed in closed form by stating a simple consistency argument (inspired by \cite{Guex-2016}),
\begin{equation}
 \Pext = \kbordermatrix{
               &          1 &  \{ 2, \dots, (n-1) \} = \mathcal{V}                                     &         n  \cr
1              &          0 & \boldsymbol{\sigma}_{\mathrm{in}}^{\text{T}} &         0  \cr
 \{ 2, \dots, (n-1) \} = \mathcal{V}  & \mathbf{0} & (\mathbf{I} - \mathbf{Diag}(\boldsymbol{\alpha})) \mathbf{P}                                           & \boldsymbol{\alpha} \cr
n              &          0 & \phantom{{}^{\text{T}}} \mathbf{0}^{\text{T}}        &         0  \cr
} \label{Eq_extendedTransitionMatrix01}
\end{equation}
where the quantity $\boldsymbol{\alpha}$ is defined in Equation (\ref{Eq_computing_alpha_matrix_form01}).
Interestingly, for nodes not belonging to $\mathcal{I}n$, these transition probabilities remain exactly the same as before, as long as the random walker is not absorbed. For the source node 1, its transition probabilities pointing to nodes in $\mathcal{I}n$ are set to $\boldsymbol{\sigma}_{\mathrm{in}}$ in order to satisfy the first constraint in Equation (\ref{Eq_flow_constraints01}).
If we do not have a good reason for choosing the weights $\mathbf{w}$, it seems reasonable to compute the consistent transition probabilities because it removes the arbitrariness associated with the choice of these weights.

\section{Optimal transport on a graph from a constrained randomized shortest paths framework}
\label{Sec_optimal_transport01}

Our formulation of the problem is based on the RSP framework defining dissimilarity measures interpolating between the shortest-path distance and the commute-time distance \cite{Kivimaki-2012,Saerens-2008,Yen-08K}. This formalism is based on full paths instead of standard ``local" flows \cite{Ahuja-1993}, and was initially inspired by a model developed in transportation science \cite{Akamatsu-1996}.

We start by providing a brief description (closely following \cite{Courtain-2020,Leleux-2021}) of the RSP formalism before defining the problem and then deriving the algorithm for solving the constraints-based multi-inputs multi-outputs transport problem on the graph $G_{\mathrm{ext}}$.

\subsection{The standard randomized shortest paths formalism}
\label{Sec_randomized_shortest_paths01}

Let us start with a short reminder about the RSP model. For the sake of simplicity, in this section, all quantities are discussed in the context of a graph $G$ with a single input node $1$ and a single target node $n$ in order to avoid more cumbersome notations.
The main idea behind the standard RSP is the following. We consider the set of all paths, or walks, $\wp \in \mathcal{P}_{1n}$ from node $1$ to absorbing node $n$ on $G$. Each path $\wp$ consists in a sequence of connected nodes starting in node $1$ and ending in $n$. Then, we assign a probability distribution $\text{P}(\cdot)$ on the set of paths $\mathcal{P}_{1n}$ by minimizing the free energy of statistical physics \cite{Jaynes-1957,Peliti-2011,Reichl-1998},
\begin{equation}
\vline\,\begin{array}{llll}
\minimize\limits_{\{ \text{P}(\wp) \}_{\wp \in \mathcal{P}_{1n}}} & \phi(\text{P}) = \dsum_{\wp \in \mathcal{P}_{1n}} \text{P}(\wp) \tilde{c}(\wp) + T \dsum_{\wp \in \mathcal{P}_{1n}} \text{P}(\wp) \log \left( \frac{\text{P}(\wp)}{\tilde{\pi}(\wp)} \right) \\[0.5cm]
\subjectto & \sum_{\wp\in\mathcal{P}_{1n}}\textnormal{P}(\wp) = 1
\end{array}
\label{Eq_optimization_problem_BoP01}
\end{equation}
with $\tilde{c}(\wp) = \sum_{\tau = 1}^{\ell} c_{\wp(\tau-1) \wp(\tau)}$ is the total cumulated cost\footnote{The basic quantities that are defined on whole paths $\wp$ will be denoted with a tilde in order to distinguish them from the same local quantities defined on edges.} along path $\wp$ when visiting the sequence of nodes (path) $\wp = \left( \wp(\tau) \right)_{\tau=0}^{\ell(\wp)}$ in the sequential order $\tau = 0,1,2,\dots,\ell(\wp)$ where $\ell(\wp)$ (or simply $\ell$) is the length of path $\wp$. Here, $ \wp(\tau)$ is the node appearing at position $\tau$ on path $\wp$. Furthermore, $\tilde{\pi}(\wp) = \prod_{\tau = 1}^{\ell} p_{\wp(\tau-1) \wp(\tau)}$ is the product of the natural random walk transition probabilities (see Equation (\ref{Eq_transition_probabilities_original_graph01})) along path $\wp$, called the path likelihood.

The objective function (\ref{Eq_optimization_problem_BoP01}) is a mixture of two dissimilarity terms with the temperature $T$ balancing the trade-off between them.
The first term is the expected cost for reaching the target node from the source node (favoring shorter paths). The second term corresponds to the relative entropy, or Kullback-Leibler divergence, between the path probability distribution and the path likelihood distribution (introducing randomness). When the temperature $T$ is low, shorter paths are favored while when $T$ is large, paths are chosen according to their likelihood in the random walk on the graph $G$. Note that we should normally add non-negativity constraints but this is not necessary as the resulting probabilities will automatically be non-negative.

This argument, akin to maximum entropy \cite{Jaynes-1957}, leads to a \textbf{Gibbs-Boltzmann distribution} on the set of paths (see, e.g., \cite{Francoisse-2017} for a detailed derivation),
\begin{equation}
\text{P}^{*}(\wp) 
= \frac{\tilde{\pi}(\wp) \exp[-\theta \tilde{c}(\wp)]}{\dsum_{\wp'\in\mathcal{P}_{1n}} \tilde{\pi} (\wp')\exp[-\theta \tilde{c}(\wp')]}
= \frac{\tilde{\pi}(\wp) \exp[-\theta \tilde{c}(\wp)]}{\mathcal{Z}}
\label{Eq_Boltzmann_probability_distribution01}
\end{equation}
where $\theta = 1/T$ is the inverse temperature and the denominator $\mathcal{Z} = \sum_{\wp\in\mathcal{P}_{1n}} \tilde{\pi} (\wp)\exp[-\theta \tilde{c}(\wp)]$ is the \textbf{partition function} of the system of paths.
It can be shown that this set of path probabilities (the randomized policy in terms of paths) is exactly generated by, and thus equivalent to, a Markov chain with biased transition probabilities (the randomized policy in terms of local transitions) favoring shorter paths, depending on the temperature $T$ (see Equation (\ref{Eq_biased_transition_probabilities01}) in the Appendix \ref{Appendix:RSP},  or \cite{Saerens-2008}).

Interestingly, if we replace the probability distribution $\text{P}(\cdot)$ by the optimal distribution $\text{P}^{*}(\cdot)$ provided by Equation (\ref{Eq_Boltzmann_probability_distribution01}) in the objective function (\ref{Eq_optimization_problem_BoP01}), we obtain
\begin{align}
\phi_{1n} \triangleq \phi(\text{P}^{*}) &= \dsum_{\wp \in \mathcal{P}_{1n}} \text{P}^{*}(\wp) \tilde{c}(\wp) + T \dsum_{\wp \in \mathcal{P}_{1n}} \text{P}^{*}(\wp) \log \left( \frac{\text{P}^{*}(\wp)}{\tilde{\pi}(\wp)} \right) \nonumber \\
 &= \dsum_{\wp \in \mathcal{P}_{1n}} \text{P}^{*}(\wp) \tilde{c}(\wp) + T \dsum_{\wp \in \mathcal{P}_{1n}} \text{P}^{*}(\wp) \left( -\tfrac{1}{T} \tilde{c}(\wp) - \log \mathcal{Z} \right) \nonumber \\
 &= -T \log \mathcal{Z}
 \label{Eq_optimal_free_energy01}
\end{align}

Furthermore, the Appendix \ref{Appendix:RSP} provides a brief summary of the most important quantities that can be derived from the standard RSP model.

\subsection{Statement of the problem}
\label{Subec_problem_statement01}

The objective now is to compute the randomized shortest paths solution on $G_{\mathrm{ext}}$ satisfying the source/target flow constraints\footnote{Required input flow $\sigma_{i}^{\mathrm{in}}$ is set to $0$ for nodes $i$ not in $\mathcal{I}n$ and output flow $\sigma_{j}^{\mathrm{out}}$ is set to $0$ for nodes $j$ not in $\mathcal{O}ut$.} stated in Section \ref{Sec_notation_problem_statement01}, and recalled here for convenience,
\begin{equation}
\begin{cases}
\bar{n}_{1i} = \sigma_{i}^{\mathrm{in}}  &\text{for each node } i \in \mathcal{I}n \\
\bar{n}_{jn} = \sigma_{j}^{\mathrm{out}} &\text{for each node } j \in \mathcal{O}ut
\end{cases}
\label{Eq_equality_constraints01}
\end{equation}
where $\bar{n}_{ij}$ is the flow (expected number of passages) in edge $(i,j)$ (see Equation (\ref{Eq_computation_edge_flows01})).

As already mentioned in the introduction, the margin-constrained BoP problem on a graph has recently been studied in \cite{Guex-2016} and \cite{Lebichot-2018}. The former paper is based on an entropy regularization at the flow level while the later one adopts a bag-of-paths approach\footnote{Observe the difference with the standard RSP formulation of Equation (\ref{Eq_optimization_problem_BoP01}).}
\begin{equation}
\vline \begin{array}{lll@{}lll}
\underset{\{\mathrm{P}(\wp) \}_{\wp \in \mathcal{P}}} {\text{minimize}} & \phi(\mathrm{P}) =  \displaystyle\sum\limits_{\wp \in \mathcal{P}} \mathrm{P}(\wp) \tilde{c}(\wp) + T \sum_{\wp \in \mathcal{P}} \mathrm{P}(\wp) \log \left( \frac{\mathrm{P}(\wp)}{\tilde{\pi}(\wp)} \right) \\
\text{subject to} & \sum_{j \in \mathcal{V}} \sum_{\wp_{ij} \in \mathcal{P}_{ij}} \mathrm{P}(\wp_{ij}) = \sigma^\mathrm{in}_i &\forall i \in \mathcal{I}n \\
          & \sum_{i \in \mathcal{V}} \sum_{\wp_{ij} \in \mathcal{P}_{ij}} \mathrm{P}(\wp_{ij}) = \sigma^\mathrm{out}_j &\forall j \in \mathcal{O}ut \\ 
\end{array}
\label{Eq_previous_formulation01}
\end{equation}
where the set of considered paths is $\mathcal{P} = \cup_{i \in \mathcal{I}n} \cup_{j \in \mathcal{O}ut} \mathcal{P}_{ij}$. 
It was shown that the solution can be obtained by iterative proportional fitting (also called matrix balancing or biproportional scaling), as for the standard, relaxed, optimal transport problem with entropy regularization (see for instance \cite{Cuturi2013,Erlander-1990,Kapur-1992,Wilson-1970}).

In the present paper, we thus adopt a different point of view (in comparison with \cite{Guex-2019}) by designing a new algorithm based on the constraints imposed on flows in the graph $G_{\mathrm{ext}}$ (see Equation (\ref{Eq_equality_constraints01})), inspired by \cite{Courtain-2020}. It is important to note that this formulation of the problem can readily integrate additional capacity constraints as well; see \cite{Courtain-2020} for details. This is what makes the present formulation interesting for practical problems: it extends the scope of the optimal transport on a graph procedure introduced in \cite{Guex-2016,Guex-2019} and solving (\ref{Eq_previous_formulation01}) to problems with capacity constraints.

\subsection{The margin-constrained randomized shortest path algorithm}
\label{Subec_optimal_transport01} 

The algorithm solving the relative entropy-regularized optimal transport on a graph problem is derived from results obtained in \cite{Courtain-2020} by exploiting its Lagrange formulation and Lagrangian duality. It provides the optimal \textbf{randomized policy} taking the form of the transition matrix of a \textbf{biased random walk} on $G_{\mathrm{ext}}$, as in the case of the standard RSP problem (see Equation \ref{Eq_biased_transition_probabilities01}). It corresponds to a Markov chain on the extended graph biasing the random walk towards the output nodes while satisfying the input and output flow constraints of Equation (\ref{Eq_equality_constraints01}).
Note that, for convenience, most of the more technical results are reported in the Appendix \ref{Sec_derivation_of_the_algorithm01}.

\subsubsection{The Lagrange function}
\label{Subsec_Lagrange_function_edge_constraints01}

The equality constraints (\ref{Eq_equality_constraints01}) can be expressed in the following Lagrange function defined on the extended graph $G_{\mathrm{ext}}$ with $\mathcal{P}_{1n}$ being the set of all possible paths from source node $1$ to target node $n$,
\begin{align}
\mathscr{L}(\text{P},\boldsymbol{\lambda})
&= \underbracket[0.5pt][3pt]{ \dsum_{\wp \in \mathcal{P}_{1n}} \text{P}(\wp) \tilde{c}(\wp) + T \dsum_{\wp \in \mathcal{P}_{1n}} \text{P}(\wp) \log \left( \frac{\text{P}(\wp)}{\tilde{\pi}(\wp)} \right) }_{\text{free energy, }\phi(\text{P})}
+ \mu \bigg( \dsum_{\wp \in \mathcal{P}_{1n}} \text{P}(\wp) - 1 \bigg) \nonumber \\
&\quad + \dsum_{i \in \mathcal{I}n} \lambda_{i}^{\mathrm{in}} \big( \bar{n}_{1i} - \sigma_{i}^{\mathrm{in}} \big)
+ \dsum_{j \in \mathcal{O}ut} \lambda_{j}^{\mathrm{out}} \big( \bar{n}_{jn} - \sigma_{j}^{\mathrm{out}} \big)
\label{Eq_Lagrange_edge_flow_constraints01}
\end{align}
where vector $\boldsymbol{\lambda}$ contains the Lagrange parameters $\{ \lambda_{i}^{\mathrm{in}} \}$ and $\{ \lambda_{j}^{\mathrm{out}} \}$.
As can be seen, there is a Lagrange parameter associated with each input node ($\mathcal{I}n$) and each output node ($\mathcal{O}ut$).
Note that because we are working on the extended graph, $\tilde{\pi}(\wp)$ is the product of the $p_{ij}^{\mathrm{ext}}$ (defined in Equation (\ref{Eq_extendedTransitionMatrix01})) along path $\wp$.

\subsubsection{Exploiting Lagrangian duality}
\label{Subsec_Lagrangian_duality_edge_constraints01}

Following the same reasoning as in \cite{Courtain-2020}, we will exploit the fact that, in this formulation of the problem, the Lagrange dual function and its gradient are easy to compute\footnote{This is actually a standard result related to maximum entropy problems (see for instance \cite{Jebara-2004}).}.
Moreover, because the objective function is convex and all the equality constraints are linear, there is only one global minimum and the duality gap is zero \cite{Bertsekas-1999,Culioli-2012,Griva-2008}. We therefore use a common optimization procedure, the Arrow-Hurwicz-Uzawa algorithm \cite{Arrow-1958}), which sequentially solves the primal and increases the dual (which is concave) until convergence.
In our context, this provides the two following steps, which are iterated until convergence,
\begin{equation}
\begin{cases}
\mathscr{L}(\text{P}^{*},\boldsymbol{\lambda}) = \min\limits_{\{ \text{P}(\wp) \}_{\wp \in \mathcal{P}_{1n}}} \mathscr{L}(\text{P},\boldsymbol{\lambda}) \text{\footnotesize{, subject to} } \sum_{\wp \in \mathcal{P}_{1n}} \text{P}(\wp) = 1  & \text{\footnotesize{(compute the dual function)}} \\
\mathscr{L}(\text{P}^{*},\boldsymbol{\lambda}^{*}) = \max\limits_{\boldsymbol{\lambda}} \mathscr{L}(\text{P}^{*},\boldsymbol{\lambda}) & \text{\footnotesize{(maximize the dual function)}}
\end{cases}
\label{Eq_primal_dual_lagrangian01}
\end{equation}
where we set $\mathscr{L}(\text{P},\boldsymbol{\lambda}) = \mathscr{L}(\text{P}^{*},\boldsymbol{\lambda}^{*})$ at the end of each iteration.
The dual function is computed analytically and then maximized through a block coordinate ascend in terms of the Lagrange parameters $\boldsymbol{\lambda}$.
It is shown in the Appendix \ref{Sec_derivation_of_the_algorithm01} that the dual function is
\begin{equation}
\mathscr{L}(\text{P}^{*},\boldsymbol{\lambda})
= -T \log \mathcal{Z}' - \dsum_{i \in \mathcal{I}n} \lambda_{i}^{\mathrm{in}} \sigma_{i}^{\mathrm{in}}
- \dsum_{j \in \mathcal{O}ut} \lambda_{j}^{\mathrm{out}} \sigma_{j}^{\mathrm{out}}
\label{Eq_dual_lagrangian01}
\end{equation}
where $\mathcal{Z}' = \sum_{\wp\in\mathcal{P}_{1n}} \tilde{\pi} (\wp)\exp[-\theta \tilde{c}'(\wp)]$ is the partition function (Equation (\ref{Eq_partition_function_definition01})) computed from the so-called \textbf{augmented costs} $c'_{ij}$ on $G_{\mathrm{ext}}$, which depend on the Lagrange multipliers,
\begin{equation}
c'_{ij} =
\begin{cases}
c_{ij}^{\mathrm{ext}} + \lambda_{j}^{\mathrm{in}} \\
c_{ij}^{\mathrm{ext}} + \lambda_{i}^{\mathrm{out}} \\
c_{ij}^{\mathrm{ext}}
\end{cases}
= \begin{cases}
\lambda_{j}^{\mathrm{in}} & \text{when } i = 1 \text{ and } j \in \mathcal{I}n \\
\lambda_{i}^{\mathrm{out}} & \text{when } i \in \mathcal{O}ut \text{ and } j = n \\
c_{ij}^{\mathrm{ext}} & \text{otherwise}
\end{cases}
\label{Eq_redefined_costs01}
\end{equation}

Moreover, as further shown in the Appendix \ref{Sec_derivation_of_the_algorithm01}, the maximization of the dual function provides the following Lagrange parameters updates at each iteration,
\begin{equation}
\begin{cases}
\lambda_{k}^{\mathrm{in}} = \tfrac{1}{\theta} \bigg( \log z'_{kn} - \dsum_{l \in \mathcal{I}n} \sigma_{l}^{\mathrm{in}} \log z'_{kn} \bigg) \text{ for } k \in \mathcal{I}n \\
\lambda_{l}^{\mathrm{out}} = \tfrac{1}{\theta} \Bigg( \log z'_{ln} - \log \bigg( \dfrac{\sigma_{l}^{\mathrm{out}}} {  p_{ln}^{\mathrm{ext}} } \bigg) - \dsum_{k \in \mathcal{O}ut} \sigma_{k}^{\mathrm{out}} \bigg[ \log z'_{kn} - \log \bigg( \dfrac{\sigma_{k}^{\mathrm{out}}} {  p_{kn}^{\mathrm{ext}} } \bigg) \bigg] \Bigg) \text{ for } l \in \mathcal{O}ut
\end{cases}
\label{Eq_lagrange_parameters_updates01}
\end{equation}
where the $z'_{kl}$ (element $k$, $l$ of the fundamental matrix) are computed thanks to Equation (\ref{Eq_fundamentalMatrix01}) in terms of the augmented costs and the natural random walk transition probabilities ($p_{kl}^{\mathrm{ext}}$) on the extended graph $G_{\mathrm{ext}}$.

\subsubsection{The resulting algorithm}

The resulting algorithm is presented in Algorithm \ref{Alg_optimal_transport01}.
The different steps of the procedure are the following:
\begin{itemize}
  \item Compute the extended graph $G_{\mathrm{ext}}$ (its edge costs and transition probabilities matrices) from the original graph $G$ as described in Subsection \ref{Subsec_extended_graph01} and Equations (\ref{Eq_extendedAdjacencyMatrix01}), (\ref{Eq_extendedCostMatrix01}) and (\ref{Eq_extendedTransitionMatrix01}). We now work on this extended graph.
  \item Initialize the Lagrange parameters to $0$.
  \item Iterate the following steps until convergence, first to update the quantities associated to the source nodes, and then to update the quantities associated to the target nodes:
  \begin{itemize}
  \item The elements of the fundamental matrix are computed from the current augmented costs (Equation (\ref{Eq_fundamentalMatrix01})) on $G_{\mathrm{ext}}$.
  \item The Lagrange parameters are updated (Equations (\ref{Eq_lagrange_parameters_updates01})).
  \item The augmented costs are updated (Equation (\ref{Eq_redefined_costs01})).
  \end{itemize}
  \item Compute the optimal policy (transition probabilities) from the augmented costs (depending on the Lagrange parameters, see Equation (\ref{Eq_redefined_costs01})) on $G_{\mathrm{ext}}$ thanks to Equation (\ref{Eq_biased_transition_probabilities01}).
\end{itemize}
The time complexity of the algorithm is dominated by the two systems of linear equations that need to be solved at each iteration. Therefore, it is of order $2k . O(n^3)$ where $n$ is the number of nodes and  $k$ is the number of required iterations. Note that Algorithm \ref{Alg_optimal_transport01} is closely related to Algorithm 2 presented in \cite{Guex-2019} page 102. Let us now introduce a dissimilarity measure based on this optimal transport model.

\subsection{A distance measure between nodes}
\label{Section:Distances}

In this subsection, we will derive two important quantities from the margin-constrained bag-of-paths model. These quantities are the coupling matrix and the surprisal distance measure between nodes of the graph, defined from the coupling matrix. For more details concerning these two quantities, see \cite{Guex-2019}.

\begin{algorithm}[t!]
\caption[Solving the relative entropy-regularized optimal transport on a graph problem with multiple sources and targets]
{\small{Solving the relative entropy-regularized optimal transport on a graph problem with multiple sources and targets, called the margin-constrained bag-of-paths model.}}

\algsetup{indent=2em, linenodelimiter=.}

\begin{algorithmic}[1]
\small
\REQUIRE $\,$ \\
 -- A weighted directed, strongly connected, graph $G_{\mathrm{ext}}$ containing $n$ nodes. Node $1$ is the source supernode and node $n$ the absorbing, target, supernode. The indegree of node 1 and the outdegree of node $n$ are both equal to $0$. \\
 -- The set of input nodes $\mathcal{I}n$ (only connected to the source supernode $1$) and output nodes $\mathcal{O}ut$ (only connected to the target supernode $n$).\\
 -- The $n\times n$ transition matrix $\Pext$ associated to $G_{\mathrm{ext}}$.\\
 -- The $n\times n$ non-negative cost matrix $\mathbf{C}_{\mathrm{ext}}$ associated to $G_{\mathrm{ext}}$ (see Equation (\ref{Eq_extendedCostMatrix01})). These original costs are equal to zero for edges starting in node $1$ and ending in $\mathcal{I}n$ as well as edges starting in $\mathcal{O}ut$ and ending in node $n$. \\
  -- The $n \times 1$ vectors of input flows, $\boldsymbol{\sigma}_{\mathrm{in}}$, and output flows, $\boldsymbol{\sigma}_{\mathrm{out}}$, both non-negative and summing to $1$.\\
  -- The inverse temperature parameter $\theta$.\\
 
\ENSURE $\,$ \\
 -- The $n \times n$ randomized policy defined by the transition matrix $\mathbf{P}^{*}$.\\

\STATE $\boldsymbol{\lambda}_{\mathrm{in}} \leftarrow \mathbf{0}$; $\boldsymbol{\lambda}_{\mathrm{out}} \leftarrow \mathbf{0}$ \COMMENT{initialize $n \times 1$ Lagrange parameter vectors} \\
\STATE $\mathbf{C}' \leftarrow \mathbf{C}_{\mathrm{ext}}$ \COMMENT{initialize the augmented costs matrix} \\
\REPEAT[main iteration loop]
\STATE $\mathbf{W}' \leftarrow \Pext\circ\exp[-\theta\mathbf{C}']$ \COMMENT{compute $\mathbf{W}'$ matrix (elementwise exponential and multiplication $\circ$)} \\
\STATE Solve $(\mathbf{I}-\mathbf{W}') \mathbf{z}'_{n} = \mathbf{e}_{n}$ \COMMENT{backward variables (column $n$ of the fundamental matrix $\mathbf{Z}'$) with elements $z'_{kn}$ ($n$ is fixed)} \\

\FORALL[compute Lagrange parameters associated to source nodes]{$k \in \mathcal{I}n$}
\STATE $\lambda_{k}^{\mathrm{in}} \leftarrow \tfrac{1}{\theta} \log z'_{kn}$ \COMMENT{compute Lagrange parameters} \\
\ENDFOR
\FORALL[update quantities associated to source nodes]{$k \in \mathcal{I}n$}
\STATE $\lambda_{k}^{\mathrm{in}} \leftarrow \lambda_{k}^{\mathrm{in}} - \dsum_{k' \in \mathcal{I}n} \sigma_{k'}^{\mathrm{in}} \lambda_{k'}^{\mathrm{in}}$ \COMMENT{normalize Lagrange parameters} \\
\STATE $c'_{1k} \leftarrow \lambda_{k}^{\mathrm{in}}$ \COMMENT{update augmented costs (recall that $c_{1k}^\mathrm{ext}=0$  for all $k \in \mathcal{I}n)$)} \\
\ENDFOR

\STATE $\mathbf{W}' \leftarrow \Pext \circ \exp[-\theta\mathbf{C}']$ \COMMENT{update $\mathbf{W}'$ matrix} \\
\STATE Solve $(\mathbf{I}-{\mathbf{W}'}^{\text{T}}) \mathbf{z}'_{1} = \mathbf{e}_{1}$ \COMMENT{forward variables (row $1$ of the fundamental matrix $\mathbf{Z}'$) with elements $z'_{1k}$ ($1$ is fixed)} \\

\FORALL[compute Lagrange parameters associated to target nodes]{$l \in \mathcal{O}ut$}
\STATE $\lambda_{l}^{\mathrm{out}} \leftarrow \tfrac{1}{\theta} \log z'_{1l} - \tfrac{1}{\theta} \log \bigg( \dfrac{\sigma_{l}^{\mathrm{out}}} {  p_{ln}^{\mathrm{ext}} } \bigg)$ \COMMENT{compute Lagrange parameters} \\
\ENDFOR
\FORALL[update quantities associated to target nodes]{$l \in \mathcal{O}ut$}
\STATE $\lambda_{l}^{\mathrm{out}} \leftarrow \lambda_{l}^{\mathrm{out}} - \dsum_{l' \in \mathcal{O}ut} \sigma_{l'}^{\mathrm{out}} \lambda_{l'}^{\mathrm{out}}$ \COMMENT{normalize Lagrange parameters} \\
\STATE $c'_{ln} \leftarrow \lambda_{l}^{\mathrm{out}}$ \COMMENT{update augmented costs (recall that $c_{ln}^\mathrm{ext}=0$ for all $l \in \mathcal{O}ut)$} \\
\ENDFOR

\UNTIL{convergence of $\boldsymbol{\lambda}_{\mathrm{in}}$, $\boldsymbol{\lambda}_{\mathrm{out}}$}
\STATE $\mathbf{P}^{*} \leftarrow (\mathbf{Diag}(\mathbf{z}'_{n}))^{-1} \mathbf{W}' \, \mathbf{Diag}(\mathbf{z}'_{n})$ \COMMENT{compute optimal policy}
\RETURN $\mathbf{P}^{*}$

\end{algorithmic}
\label{Alg_optimal_transport01}

\end{algorithm}

\subsubsection{The coupling matrix}

Let us first remind the definition of the \textbf{coupling matrix} \cite{villani2003topics,villani2008optimal}, which was used in \cite{Guex-2019} to define a distance measure between the nodes of the graph (see next subsection).
This coupling matrix is denoted by $\mathbf{\Gamma} = (\gamma_{ij})$ and is defined as the joint probability of starting the walk in node $i \in \mathcal{I}n$ (reaching input node $S = i$ from supernode $1$ at time step $1$) and ending the walk in node $j \in \mathcal{O}ut$ (visiting output node $T = j$ at time step $\ell(\wp) - 1$ and then immediately transiting to supernode $n$) when walking according to the optimal path probabilities defined in Equations (\ref{Eq_Boltzmann_probability_distribution01}) and (\ref{Eq_Boltzmann_probability_distribution02}) and using the augmented costs after convergence in order to satisfy the constraints,
\begin{align}
\gamma_{ij} &\triangleq \mathrm{P}^{*}(S = i, T = j)
= \mathrm{P}^{*}(\wp(1) = i, \wp(\ell - 1) = j | \wp(0) = 1, \wp(\ell) = n) \nonumber \\
&= \frac{w'_{1i} \big( \sum_{\wp_{ij} \in \mathcal{P}_{ij}} \tilde{w}'(\wp_{ij}) \big) w'_{jn}} {\sum_{\wp'\in\mathcal{P}} \tilde{w}(\wp')}
= \frac{w'_{1i} z'_{ij} w'_{jn}} {\sum_{\wp'\in\mathcal{P}} \tilde{w}'(\wp')} \nonumber \\
&= \frac{w'_{1i} z'_{ij} w'_{jn}} {\sum_{i' \in \mathcal{I}n} \sum_{j' \in \mathcal{O}ut} w'_{1i'} z'_{i'j'} w'_{j'n}} \quad \text{with } i \in \mathcal{I}n \text{ and } j \in \mathcal{O}ut
\label{Eq_coupling_probabilities01}
\end{align}
where $\mathcal{P}_{ij}$ is the set of paths starting in input node $i \in \mathcal{I}n$ and ending in output node $j \in \mathcal{O}ut$.
It also holds that $\sum_{i \in \mathcal{I}n} \gamma_{ij} = \sigma_{j}^{\mathrm{out}}$ and $\sum_{j \in \mathcal{O}ut} \gamma_{ij} = \sigma_{i}^{\mathrm{in}}$.
Notably, the elements $w'_{ij}$ are computed from the augmented costs after convergence of Algorithm \ref{Alg_optimal_transport01}. Further note that the rows and the columns associated with the two supernodes indexed as nodes $1$ and $n$ should be removed to obtain the coupling matrix associated with the original graph.

\subsubsection{The margin-constrained bag-of-paths surprisal distance}
\label{Subsection:cBoP}

We can now compute the \textbf{margin-constrained bag-of-paths surprisal distance} introduced in \cite{Guex-2019} (called the margin-constrained bag-of-paths surprisal distance in this paper, and itself inspired by \cite{Francoisse-2017}) as
\begin{equation}
\myDelta_{ij}^{\mathrm{sur}} =
\begin{cases}
-\tfrac{1}{2}(\mathrm{log}(\gamma_{ij})+\mathrm{log}(\gamma_{ji}))&\text{if } i \neq j,\\
\hspace{3.1mm} 0 & \text{if } i = j\\
\end{cases} 
\label{Eq_surprisal}
\end{equation}
where $\gamma_{ij}$ is the elements of the coupling matrix $\mathbf{\Gamma}$ defined in Equation (\ref{Eq_coupling_probabilities01}). This distance is a generalization of the surprisal distance \cite{Francoisse-2017,Kivimaki-2012} where positive weights $\mathbf{v}=(v_i)$, summing to 1, can be attached to each node through $\boldsymbol{\sigma}_{\mathrm{in}}$ and $\boldsymbol{\sigma}_{\mathrm{out}}$.

Intuitively, the distance (\ref{Eq_surprisal}) quantifies the ``surprise" generated by the event $(S = i) \land (T = j)$, that is, picking a path with input node $i \in \mathcal{I}n$ and output node $j \in \mathcal{O}ut$ from the bag of paths $\mathcal{P}_{1n}$ defined on $G_{\mathrm{ext}}$, with probability distribution (\ref{Eq_Boltzmann_probability_distribution01}) and using augmented costs in order to satisfy the constraints (\ref{Eq_equality_constraints01}).

In this work, we consider that each node acts as both input and output, with $\boldsymbol{\sigma}_{\mathrm{in}}=\boldsymbol{\sigma}_{\mathrm{out}}=\mathbf{v}$. This choice is inspired from the PageRank \cite{Brin-1998,Page-1998} and the random walk with restart \cite{Gori-2007,Tong-2007} models in which, at each time step, the random walker has a chance of leaving the current node (which is then similar to a sink node) for restarting in some nodes of the network (which are then similar to source nodes).
This choice, although somewhat counter-intuitive\footnote{Because we inject and remove the same quantity of  flow in each node, this setting is only meaningful when the parameter $\theta$ is not too large. Indeed, when $\theta$ increases, the transportation is more and more optimal and the input-output flows tend to neutralize each other, resulting in a coupling matrix that converges to the identity matrix. This is, however, not the case for intermediate measures of $\theta$ for which the coupling matrix is able to capture the similarity between nodes (in terms of proximity and high connectivity).}, provides quite competitive results as will be shown in the next section.

Note that the authors of \cite{Guex-2019} also propose a hitting path version of this distance which is not definable in our present framework. This distance based on hitting paths will nevertheless be investigated in our experiments.


\section{Experimental comparison on semi-supervised classification tasks}
\label{Sec_Experimental_Comparison01}

In this section, the introduced method of Subsection \ref{Section:Distances} and its equivalent on hitting paths \cite{Guex-2019} will be compared in terms of classification accuracy on semi-supervised classification tasks with the other methods defined in the bag-of-paths framework. The goal of this experiment is to highlight the best methods within the bag-of-paths framework rather than propose an extended comparison with a large number of state-of-the-art techniques. Indeed, the methods defined in the bag-of-paths framework, like the free energy distance, have already demonstrated their competitiveness with state-of-the-art techniques in some pattern recognition tasks \cite{Francoisse-2017,Guex-2021,Sommer-2016}.  

The section is organized as follows. First, the set of investigated methods is presented in Subsection \ref{SubSection:Methods}. Then, Subsection \ref{SubSection:ExpDesign} provides details on the experimental design inspired by \cite{Courtain-2020,Francoisse-2017}. Finally, Subsection \ref{SubSection:Results} presents and discusses the results of the experiments.

\subsection{Investigated methods}
\label{SubSection:Methods}

For our experimental comparisons, we have selected seven methods defined in the RSP/BoP framework, introduced in Subsection \ref{Sec_randomized_shortest_paths01}. Recall that the main difference between these two models is that the randomized shortest paths model is defined based on the set of all (usually hitting) paths $\mathcal{P}_{st}^{\mathrm{h}}$ from a unique node $s$ to a unique target node $t$, and not based on the set of all paths $\mathcal{P}$ in the graph, as the bag-of-paths model does. Nevertheless, most of the methods introduced hereafter could be defined in both frameworks.

Moreover, as already stated in other terms in Subsection \ref{Subsec_problem_statement01}, the RSP framework interpolates between an optimal (exploitation) and a random (exploration) behavior based on a monitoring parameter $\theta$. This framework therefore allowed us to define dissimilarity measures interpolating between the shortest path distance when $\theta$ is large (exploitation) and the commute time distance (up to a scaling factor) when $\theta \rightarrow 0^{+}$ (exploration). Therefore, it seemed clear to us to use these two boundary methods as baselines in our experiments. Finally, the seven methods retained for the experiments are:

\begin{itemize}
    \item The \emph{shortest path distance} (SP) between two nodes $i$ and $j$ is defined as the path with the smallest cost between these two nodes, derived from the cost matrix $\mathbf C$. This method is the most standard distance and has no hyperparameter.
    \item The \emph{commute time kernel} (CT) \cite{FoussKDE-2005,Saerens04PCA} simply corresponds to the Moore-Penrose pseudoinverse of the Laplacian matrix $\mathbf{L^+}$ \cite{Fouss-2016}. This method has no hyperparameter.
    \item The \emph{free energy distance} (FE) \cite{Francoisse-2017,Kivimaki-2012} is a distance build on the symmetrization of the directed free energy distance presented in Equation (\ref{Eq_optimal_free_energy01}). This method has one hyperparameter $\theta$.
    \item The regular \emph{surprisal distance} (Sup) \cite{Francoisse-2017} is a distance quantifying the ``surprise" generated by the event $(S = i) \land (T = j)$ (see Subsection \ref{Subsection:cBoP}). This method has one hyperparameter $\theta$.
    \item The regular \emph{randomized shortest paths dissimilarity} (RSP) \cite{Kivimaki-2012,Saerens-2008} is obtained by symmetrization of the expected cost of Equation (\ref{Eq_real_expected_cost01}). This method has one hyperparameter $\theta$.
    \item The \emph{margin-constrained bag-of-paths surprisal distance} (cBoP) is the distance introduced in \cite{Guex-2019} and re-derived in this paper (Subsection \ref{Subsection:cBoP}) from another point of view. This method has two hyperparameters $\theta$ and the non-negative weights vector $\mathbf{v}$.
    \item The \emph{margin-constrained bag-of-hitting-paths surprisal distance} (cBoPH) \cite{Guex-2019} is the counterpart of the previous method in terms of hitting paths. This method has two hyperparameters $\theta$ and the non-negative weights vector $\mathbf{v}$. We computed the quantity by following \cite{Guex-2019}, Algorithm 3, page 108.
\end{itemize}

\subsection{Experimental design}
\label{SubSection:ExpDesign}

As mentioned earlier, our experimental methodology is closely related to the one used in \cite{Courtain-2020,Francoisse-2017}. This design was initially inspired by \cite{Tang-2009,Tang-2009b,Tang-2010,Zhang-2008b,Zhang-2008} and was recently used in \cite{Guex-2021} leading to interesting results. Therefore, the following section will only summarize the procedure\footnote{The interested reader can found a more complete description in Section 7 of \cite{Francoisse-2017}.} and emphasize the main differences.

\subsubsection{Datasets}

A collection of 14 well-known network datasets, already used in previous experimental comparisons, has been selected to evaluate the performance of the different methods. The collection includes the 4 WebKB datasets \cite{Macskassy-07}, the IMDB dataset \cite{Macskassy-07}, and 9 extracted from the 20 Newsgroup datasets \cite{lichman2013uci,Yen-2009}. All these datasets are described by an adjacency matrix \textbf{A} and a class label vector \textbf{y}. Note that we consider that each graph is undirected and we assert it by using $\mathbf{A} = (\mathbf{A}+\mathbf{A}^{\mathrm{T}})/2$. Furthermore, the elements of the cost matrix $\mathbf{C}$ are defined as $c_{ij} = 1/a_{ij}$ as for electrical networks \cite{Francoisse-2017}. A summary of the main characteristics of each dataset can be found in Table \ref{Table:datasets}.

\begin{table}[t]
\centering
\footnotesize
\begin{tabular}{|l|c|c|c|c|}
\hline
\textbf{Dataset Name} & \mypound \textbf{Labels} & \mypound \textbf{Nodes} & \mypound \textbf{Edges} & \textbf{Prior of the majority class} \\ \hline
webKB-cornell (DB1)  & 6        & 346   & 13416 & 41.91\% \\ \hline
webKB-texas  (DB2)& 6        & 334   & 16494 & 48.80\% \\ \hline
webKB-washington (DB3) & 6        & 434   & 15231 & 39.17\% \\ \hline
webKB-wisconsin (DB4)& 6        & 348   & 16625 & 44.54\% \\ \hline
imdb   (DB5)          & 2        & 1126  & 20282 & 50.18\% \\ \hline       
news-2cl-1 (DB6) & 2        & 400   & 33854 & 50.00\% \\ \hline
news-2cl-2 (DB7) & 2        & 398   & 21480 & 50.25\% \\ \hline
news-2cl-3 (DB8) & 2        & 399   & 36527 & 50.13\% \\ \hline
news-3cl-1 (DB9)& 3        & 600   & 70591 & 33.34\% \\ \hline
news-3cl-2 (DB10)& 3        & 598   & 68201 & 33.44\% \\ \hline
news-3cl-3 (DB11)& 3        & 595   & 64169 & 33.61\%  \\ \hline
news-5cl-1 (DB12)& 5        & 998   & 176962 & 20.04\%\\ \hline
news-5cl-2 (DB13) & 5        & 999   & 164452 & 20.02\% \\ \hline
news-5cl-3 (DB14) & 5        & 997   & 155618 & 20.06\% \\ \hline
\end{tabular}
\caption{Main characteristics of the datasets used in our experiments.}
\label{Table:datasets}
\end{table}

\subsubsection{Experimental methodology}

The graph-based semi-supervised classification methodology is divided into two parts. The first part consists of extracting $\{5\%,10\%,20\%\}$ of the dominant eigenvectors of a kernel matrix to use them as node features in a linear support vector machine (SVM)\footnote{We use the LIBSVM library \cite{LIBSVM} with the options '-s 0' and '-t 0'.}. These extracted features contain condensed information about the graph structure.

The second part consists of directly feed the kernel matrix into a kernel SVM\footnote{We use the LIBSVM library \cite{LIBSVM} with the options '-s 0' and '-t 4'.}. We will refer to the first part as 5\%F, 10\%F, and 20\%F and the second part as Ker. The main objective is to determine to which extent the different methods can deal with partial information about the graph structure, contained in only a few dimensions ($5\%$, $10\%$ and $20\%$ -- node features extraction), as well as with the full information contained in the kernel matrix.

The kernel matrices $\mathbf{K}$ are derived from the dissimilarity matrices by using both classical multidimensional scaling (MDS) \cite{Borg-1997} and Gaussian transformation (Gauss) \cite{Scholkopf-2002}. We also considered centering the Gaussian kernels (GaussCenter)\footnote{We did not apply this transformation to the MDS kernels as they are centered by construction.} by applying the following transformation $\mathbf{K}=\mathbf{H}\mathbf{K}\mathbf{H}$ where $\mathbf{H} = \mathbf{I} - \mathbf{e} \mathbf{e}^{\mathrm{T}}/n$ is the centering matrix, $\mathbf{e}$ is a column vector full of 1's and $n$ is the number of nodes.

For the first part of the classification method, we also try two different options for extracting the node features. The first option is to weight the dominant eigenvectors by the square root of their corresponding eigenvalues before concatenating them into the data matrix $\mathbf{X}$. The matrix $\mathbf{X}$ contains the features of the nodes on its rows and is used as the input of the SVM. This option is equivalent to a multidimensional scaling limited to a reduced number of dimensions and it is denoted as SD (spectral decomposition). The second option corresponds to directly concatenating the dominant eigenvectors into the matrix $\mathbf{X}$ and to normalize each row, in such a way that the resulting node feature vectors are of unit length. This normalization corresponds to a projection of the rows of $\mathbf{X}$ on the unit radius sphere centered at the origin that removes the effect of the size of the feature vectors (only the direction is relevant). We will refer to this second option as NSD (normalized spectral decomposition). For conciseness, we only report for each method the results of the best kernel transformation and feature extraction options according to the final Nemenyi tests \cite{Demsar-2006}. The best combination for each method is reported in Table \ref{Tab:CombiKernelFE}.

The performance of the different methods will be evaluated in terms of classification accuracy. To avoid large variance in the results, all the methods are assessed by repeating 10 runs of a standard $5 \times 5$ nested cross-validation methodology with different folds of labeled/unlabeled nodes. In each external 5-folds cross-validation, methods are trained on 1 fold containing $20\%$ of the labels, and the remaining $80\%$ of the labels are hidden. The parameters of each method are tuned on the training fold by performing an internal 5-fold cross-validation with a labeling rate of $80\%$. In each run, external and internal folds are kept identical for all methods. The final accuracy and standard deviation are obtained by averaging the 50 results of the external cross-validation folds.

Concerning the parameters, the $\theta$ of the bag-of-paths-type methods are tuned among values of $\{10^{-6},10^{-5},10^{-4},10^{-3},10^{-2},10^{-1},1,10\}$, and the margin parameter $c$ of the SVM is tuned on the set of values $\{10^{-2},10^{-1}, 1, 10, 100\}$. For the margin-constrained bag-of-paths methods introduced in \cite{Guex-2019} and reinterpreted in Subsection \ref{Subsection:cBoP}, we tested the three following different types of positive weights $\mathbf{v}$:
\begin{itemize}
    \item Uniform weights: $\mathbf{e}/n$;
    \item L1-normalized degree weights: $\mathbf{d} / (\mathbf{e}^{\text{T}}\mathbf{d})$ with $\mathbf{d} = \mathbf{A} \mathbf{e}$;
    \item L1-normalized inverse degree weights: $(\mathbf{e} \div \mathbf{d}) / \big( \mathbf{e}^{\text{T}} (\mathbf{e} \div \mathbf{d}) \big)$. 
\end{itemize}
where $\div$ is the elementwise division.
To avoid redundancy in the comparisons, we only present the best results over the three (positive sum-to-one) weightings for each of the two margin-constrained bag-of-paths methods according to Nemenyi tests \cite{Demsar-2006}. We observed that, for every feature extractions sets, the weight achieving the best result is the normalized degree for both hitting and non-hitting margin-constrained bag-of-paths surprisal distances.

\begin{table}[t]
\footnotesize
\begin{center}
\scalebox{1}{
\begin{tabular}{|l|c|c|c|c|}
\hline
Methods            & 5\%F         & 10\%F        & 20\%F        & Ker     \\ \hline
CT & NSD           & NSD         & NSD        & /           \\ \hline
cBoP           & GaussCenterSD & GaussCenterSD & GaussCenterSD & GaussCenter \\ \hline
cBoPH          & GaussCenterSD & GaussSD       & GaussCenterSD & Gauss       \\ \hline
FE       & GaussCenterSD & GaussCenterSD & GaussSD       & Gauss       \\ \hline
RSP                 & GaussCenterSD & GaussSD       & GaussCenterSD & Gauss       \\ \hline
SP                  & GaussSD       & GaussSD       & GaussCenterSD & GaussCenter \\ \hline
Sup          & GaussSD       & GaussSD       & GaussCenterSD & Gauss       \\ \hline
\end{tabular}}
\caption{\footnotesize{The best combination of kernel transformation and feature extraction options for each method according to Nemenyi tests.}}
\label{Tab:CombiKernelFE}
\end{center}
\end{table}

\subsection{Results and discussion}
\label{SubSection:Results}

The classification accuracy and standard deviation averaged over the 50 runs are reported in Table \ref{table:classres} for the 14 datasets and the four extracted feature sets. The best performing method is highlighted in boldface for each dataset and each feature set. Bold values highlighted in grey indicate the best performing method overall (across all feature sets) for each dataset.

\subsubsection{Comparison of the different methods}

The classification results of the seven different methods are now compared for each feature set, and then across all feature sets.

From the raw results of Table \ref{table:classres}, it can be seen that 11 of the best results over 14 datasets are obtained by directly feeding the kernel to the SVM. Furthermore, 9 of these 14 results are obtained by the two introduced constrained optimal transport methods (cBoP and cBoPH). Across all the feature sets, we observe that the cBoP seems to outperform all the methods on DB1 to DB4 except in 20\%F where the RSP prevails on DB1. The results on DB5 are more contrasted: the FE performs the best for 10\%F and 20\%F, whereas the CT and the Sup are respectively the highlighted methods for 5\%F and Ker. For the newsgroup datasets (DB6-DB14), the best method is dataset-dependent and feature set-dependent except for the DB8 where the SP dominates. Nevertheless, we can underline that the cBoPH appears 18 times among the 32 highlighted results of the eight other newsgroup datasets. The remaining methods only appear respectively six times for the FE, five times for the Sup, two times for the RSP, and one time for the CT. Another observation is that the CT obtains slightly lower results with Ker compared to its performances in the other feature sets. The additional information provided by the kernel is therefore not optimally exploited by this method.

\setBoldness{0.5}%
\begin{table}[t!]
\footnotesize
\begin{center}
\scalebox{0.75}{
\begin{tabular}{lccccccc}
\hline
\textbf{Method $\rightarrow$}: & \multicolumn{1}{c}{\multirow{2}{*}{CT}} & \multicolumn{1}{c}{\multirow{2}{*}{cBoP}} & \multicolumn{1}{c}{\multirow{2}{*}{cBoPH}} & \multicolumn{1}{c}{\multirow{2}{*}{FE}} & \multicolumn{1}{c}{\multirow{2}{*}{RSP}} & \multicolumn{1}{c}{\multirow{2}{*}{SP}} & \multicolumn{1}{c}{\multirow{2}{*}{Sup}} \\
\textbf{Dataset} $\downarrow$:& \multicolumn{1}{c}{}&\multicolumn{1}{c}{}   &\multicolumn{1}{c}{}   &\multicolumn{1}{c}{}   &\multicolumn{1}{c}{}  &\multicolumn{1}{c}{}  & \multicolumn{1}{c}{}      \\
\hline
 &\multicolumn{6}{c}{\textbf{5\%F}}  \\
\hline
webKB-cornell &54.10$\pm$3.91 & {\fbseries59.41$\pm$3.88} & 58.82$\pm$2.82 & 59.07$\pm$2.85 & 58.82$\pm$2.90 & 47.65$\pm$3.43 & 58.67$\pm$3.15 \\
webKB-texas & 67.24$\pm$3.40 & {\fbseries78.62$\pm$2.34} & 77.44$\pm$3.05 & 77.87$\pm$3.13 & 76.28$\pm$3.52 & 66.89$\pm$2.83 & 77.14$\pm$2.94 \\
webKB-washington & 65.59$\pm$2.10 &{\fbseries71.10$\pm$3.02} & 69.65$\pm$2.74 & 69.50$\pm$2.83 & 69.49$\pm$2.87 & 62.75$\pm$2.69 & 68.98$\pm$2.61 \\
webKB-wisconsin &71.88$\pm$3.53 &{\fbseries79.43$\pm$1.99} & 78.53$\pm$2.07 & 78.16$\pm$1.87 & 76.62$\pm$2.23 & 64.14$\pm$3.12 & 77.67$\pm$2.24 \\
imdb & {\fbseries78.90$\pm$1.31} &77.61$\pm$1.20& 78.29$\pm$1.74 & 78.34$\pm$1.76 & 78.09$\pm$1.45 & 77.58$\pm$1.57 & 78.28$\pm$1.42 \\
news-2cl-1 &\cellcolor[HTML]{C0C0C0}{\fbseries96.64$\pm$0.73} & 95.50$\pm$1.30 & 95.84$\pm$1.04 & 95.66$\pm$1.18 & 95.56$\pm$1.74 & 92.82$\pm$1.30 & 95.59$\pm$1.30 \\
news-2cl-2 &91.17$\pm$1.34 & 91.64$\pm$2.22 & 93.29$\pm$1.50 & 93.22$\pm$1.41 & 92.54$\pm$1.96 & 91.61$\pm$1.81 & \cellcolor[HTML]{C0C0C0}{\fbseries93.47$\pm$1.40} \\
news-2cl-3 &94.08$\pm$0.68 & 96.18$\pm$1.01 & 96.47$\pm$0.81 & 96.39$\pm$0.90 & 96.19$\pm$1.26 &{\fbseries96.59$\pm$0.86} & 96.40$\pm$1.02 \\
news-3cl-1 &90.52$\pm$1.32 & 92.66$\pm$1.28 & {\fbseries93.05$\pm$1.43}& 92.82$\pm$1.48 & 92.75$\pm$1.48 & 92.87$\pm$1.18 & 92.91$\pm$1.38 \\
news-3cl-2 &89.92$\pm$1.23 & 92.00$\pm$1.51 & {\fbseries92.50$\pm$1.38} & 92.29$\pm$1.39 & 92.07$\pm$1.43 & 89.61$\pm$1.46 & 92.37$\pm$1.66 \\
news-3cl-3 &89.63$\pm$1.21 & 92.14$\pm$1.65 & {\fbseries92.91$\pm$1.36} & 92.47$\pm$1.61 & 92.06$\pm$1.81 & 91.42$\pm$1.00 & 92.71$\pm$1.40 \\
news-5cl-1 &78.50$\pm$1.55 & 88.49$\pm$0.95 & {\fbseries88.79$\pm$1.05} & 88.49$\pm$1.17 & 88.43$\pm$1.05 & 87.93$\pm$1.09 & 88.49$\pm$1.37 \\
news-5cl-2 &76.04$\pm$1.61 & 82.76$\pm$1.42 & 83.23$\pm$1.33 & 83.33$\pm$1.40 & 82.58$\pm$1.76 & 81.80$\pm$1.24 & {\fbseries83.35$\pm$1.37} \\
news-5cl-3 &75.25$\pm$1.78 & 82.19$\pm$1.36 & 82.51$\pm$1.73 & {\fbseries82.81$\pm$1.80} & 82.31$\pm$1.53 & 77.17$\pm$1.77 & 82.49$\pm$1.62 \\
\hline 
 &\multicolumn{6}{c}{\textbf{10\%F}}  \\
\hline
webKB-cornell & 53.51$\pm$3.28 & {\fbseries58.08$\pm$3.25} & 57.98$\pm$3.19 & 57.79$\pm$3.10 & 57.95$\pm$3.95 & 48.29$\pm$3.27 & 57.00$\pm$3.89 \\
webKB-texas & 70.26$\pm$2.71 & {\fbseries78.05$\pm$3.14} & 76.82$\pm$3.03 & 76.81$\pm$3.27 & 76.48$\pm$2.49 & 65.86$\pm$3.25 & 76.04$\pm$3.46 \\
webKB-washington & 63.53$\pm$2.56 & {\fbseries70.46$\pm$2.58} & 68.29$\pm$2.46 & 68.63$\pm$2.15 & 69.59$\pm$2.78 & 62.40$\pm$2.69 & 67.75$\pm$2.54 \\
webKB-wisconsin & 73.00$\pm$1.36 & \cellcolor[HTML]{C0C0C0}{\fbseries79.61$\pm$2.01} & 77.67$\pm$2.17 & 77.59$\pm$1.97 & 77.06$\pm$2.05 & 63.79$\pm$2.59 & 76.52$\pm$2.11 \\
imdb & 77.35$\pm$1.31 & 77.76$\pm$1.31 & 78.95$\pm$1.42 & {\fbseries79.10$\pm$1.38} & 78.83$\pm$1.36 & 77.57$\pm$1.56 & 78.63$\pm$1.55 \\
news-2cl-1 &93.30$\pm$1.64 & 95.00$\pm$1.06 & 95.08$\pm$1.69 & 95.06$\pm$1.30 &{\fbseries 95.17$\pm$1.43} & 93.15$\pm$1.19 & 94.90$\pm$1.73 \\
news-2cl-2 &91.79$\pm$1.79 & 91.76$\pm$2.04 & 92.36$\pm$1.69 & 92.32$\pm$1.64 & 91.87$\pm$1.73 & 90.78$\pm$2.17 & {\fbseries92.41$\pm$1.52} \\
news-2cl-3 &93.53$\pm$1.17 & 96.22$\pm$0.83 & 96.17$\pm$1.13 & 96.29$\pm$1.02 & 96.36$\pm$1.03 & {\fbseries96.72$\pm$0.74} & 96.28$\pm$1.14 \\
news-3cl-1 &89.84$\pm$1.26 & 92.78$\pm$1.14 & 93.02$\pm$1.66 & {\fbseries93.08$\pm$1.25} & 92.63$\pm$1.53 & 92.83$\pm$1.09 & 92.69$\pm$1.46 \\
news-3cl-2 &90.28$\pm$1.56 & 92.23$\pm$1.53 & {\fbseries92.72$\pm$1.39} & 92.23$\pm$1.46 & 92.10$\pm$1.56 & 89.72$\pm$1.55 & 92.25$\pm$1.67 \\
news-3cl-3 &90.87$\pm$1.33 & 91.97$\pm$1.64 & {\fbseries92.97$\pm$1.47} & 92.23$\pm$1.48 & 92.21$\pm$1.27 & 91.72$\pm$1.12 & 92.76$\pm$1.46 \\
news-5cl-1 &83.95$\pm$1.16 & 88.56$\pm$1.03 & 88.79$\pm$1.26 & 88.71$\pm$0.96 & 88.45$\pm$1.30 & 88.01$\pm$0.81 & {\fbseries88.89$\pm$1.07} \\
news-5cl-2 &77.20$\pm$1.74 & 82.70$\pm$1.86 & {\fbseries83.56$\pm$1.52} & 83.18$\pm$2.02 & 82.42$\pm$1.96 & 81.55$\pm$1.35 & 83.36$\pm$1.88 \\
news-5cl-3 &80.50$\pm$1.82 & 82.16$\pm$1.82 & {\fbseries82.79$\pm$1.66} & 82.70$\pm$1.96 & 82.58$\pm$1.63 & 77.52$\pm$1.80 & 82.59$\pm$1.71 \\
\hline 
 &\multicolumn{6}{c}{\textbf{20\%F}}  \\
\hline
webKB-cornell    & 56.19$\pm$2.87 & 57.72$\pm$3.44 & 57.59$\pm$2.75 & 57.81$\pm$3.64 & {\fbseries58.13$\pm$3.71} & 48.46$\pm$3.62 & 56.54$\pm$3.32 \\
webKB-texas      & 75.18$\pm$1.91 & {\fbseries77.66$\pm$3.48} & 75.91$\pm$3.95 & 76.72$\pm$3.60 & 76.45$\pm$3.22 & 65.06$\pm$3.32 & 76.12$\pm$3.57 \\
webKB-washington & 65.59$\pm$1.66 & {\fbseries70.41$\pm$2.31} & 68.95$\pm$2.23 & 69.36$\pm$2.04 & 69.83$\pm$2.80 & 63.12$\pm$2.31 & 67.61$\pm$2.31 \\
webKB-wisconsin  & 74.48$\pm$1.71 & {\fbseries79.21$\pm$2.12} & 76.98$\pm$2.15 & 78.09$\pm$2.69 & 77.00$\pm$2.28 & 63.04$\pm$2.83 & 75.57$\pm$2.46 \\
imdb             & 78.38$\pm$1.48 & 77.75$\pm$1.47 & 78.83$\pm$1.84 & {\fbseries79.06$\pm$1.51} & 78.71$\pm$1.66 & 77.43$\pm$1.69 & 78.42$\pm$1.54 \\
news-2cl-1       & 87.15$\pm$2.29 & 94.76$\pm$1.44 & {\fbseries95.16$\pm$1.37} & 95.00$\pm$1.53 & 95.00$\pm$2.07 & 93.31$\pm$1.13 & 94.95$\pm$1.67 \\
news-2cl-2       & 86.75$\pm$2.48 & 91.31$\pm$2.14 & 91.92$\pm$1.88 & {\fbseries92.01$\pm$1.41} & 91.89$\pm$1.53 & 90.94$\pm$1.81 & 91.63$\pm$2.17 \\
news-2cl-3       & 90.46$\pm$1.53 & 96.29$\pm$0.88 & 96.35$\pm$0.93 & 96.30$\pm$0.98 & 96.29$\pm$1.10 & {\fbseries96.64$\pm$1.00} & 96.38$\pm$0.97 \\
news-3cl-1       & 84.03$\pm$1.95 & 92.69$\pm$1.39 & 93.00$\pm$1.25 & {\fbseries93.01$\pm$1.40} & 92.50$\pm$1.43 & 92.82$\pm$0.98 & 92.99$\pm$1.27 \\
news-3cl-2       & 85.74$\pm$1.89 & 92.32$\pm$1.21 & {\fbseries92.99$\pm$1.22} & 92.69$\pm$1.32 & 92.69$\pm$1.08 & 89.57$\pm$1.39 & 92.81$\pm$1.41 \\
news-3cl-3       & 87.42$\pm$2.06 & 92.15$\pm$1.44 & 92.71$\pm$1.19 & {\fbseries92.80$\pm$1.10} & 92.38$\pm$0.94 & 91.82$\pm$1.14 & 92.80$\pm$1.16 \\
news-5cl-1       & 81.21$\pm$1.66 & 88.55$\pm$1.12 & {\fbseries88.91$\pm$1.07} & 88.75$\pm$0.98 & 88.53$\pm$1.15 & 88.09$\pm$0.93 & 88.84$\pm$1.12 \\
news-5cl-2       & 75.69$\pm$1.82 & 82.85$\pm$1.48 & {\fbseries83.48$\pm$1.52} & 83.25$\pm$1.45 & 82.27$\pm$1.92 & 81.69$\pm$1.46 & 83.44$\pm$1.41 \\
news-5cl-3       & 75.81$\pm$1.84 & 82.42$\pm$1.57 & 82.81$\pm$1.80 & {\fbseries82.94$\pm$1.61} & 81.98$\pm$1.92 & 77.57$\pm$1.49 & 82.86$\pm$1.80 \\
\hline 
 &\multicolumn{6}{c}{\textbf{Ker}}  \\
\hline
webKB-cornell    & 42.05$\pm$0.40 & \cellcolor[HTML]{C0C0C0}{\fbseries59.45$\pm$2.99} & 58.79$\pm$3.38 & 58.51$\pm$2.81 & 58.30$\pm$3.76 & 48.58$\pm$3.68 & 58.71$\pm$2.75 \\
webKB-texas      & 50.13$\pm$2.04 & \cellcolor[HTML]{C0C0C0}{\fbseries78.87$\pm$2.81} & 76.68$\pm$3.26 & 77.11$\pm$3.17 & 76.88$\pm$2.93 & 64.50$\pm$3.57 & 76.91$\pm$3.05 \\
webKB-washington & 44.44$\pm$5.54 & \cellcolor[HTML]{C0C0C0}{\fbseries71.84$\pm$2.04} & 70.12$\pm$2.19 & 69.46$\pm$2.02 & 69.78$\pm$3.14 & 63.08$\pm$2.55 & 69.05$\pm$2.20 \\
webKB-wisconsin  & 53.28$\pm$4.26 & {\fbseries78.35$\pm$1.97} & 76.42$\pm$2.42 & 77.26$\pm$2.24 & 75.69$\pm$2.32 & 63.96$\pm$2.93 & 76.25$\pm$2.32 \\
imdb             & 79.15$\pm$1.08 & 77.86$\pm$1.57 & 79.24$\pm$1.48 & 78.95$\pm$1.51 & 78.99$\pm$1.21 & 77.57$\pm$1.97 & \cellcolor[HTML]{C0C0C0}{\fbseries79.58$\pm$1.52} \\
news-2cl-1       & 90.21$\pm$7.25 & 94.98$\pm$1.28 & 95.63$\pm$0.88 & 95.40$\pm$1.21 & {\fbseries95.66$\pm$1.50} & 93.41$\pm$1.08 & 95.45$\pm$0.88 \\
news-2cl-2       & 92.08$\pm$2.03 & 91.56$\pm$1.54 & {\fbseries92.59$\pm$1.60} & 92.27$\pm$1.61 & 92.04$\pm$1.60 & 91.25$\pm$1.56 & 92.32$\pm$1.97 \\
news-2cl-3       & 87.54$\pm$9.94 & 96.34$\pm$1.06 & 96.62$\pm$0.74 & 96.62$\pm$0.78 & 96.47$\pm$0.96 & \cellcolor[HTML]{C0C0C0}{\fbseries96.79$\pm$0.78} & 96.72$\pm$0.84 \\
news-3cl-1       & 63.96$\pm$15.90 & 92.85$\pm$1.25 & \cellcolor[HTML]{C0C0C0}{\fbseries93.33$\pm$1.00} & 93.24$\pm$0.92 & 92.80$\pm$1.32 & 92.84$\pm$1.00 & 93.29$\pm$1.02 \\
news-3cl-2       & 54.96$\pm$15.96 & 92.56$\pm$1.14 & 93.23$\pm$1.00 & 92.84$\pm$1.05 & 92.65$\pm$1.16 & 89.75$\pm$1.28 & \cellcolor[HTML]{C0C0C0}{\fbseries93.26$\pm$0.97} \\
news-3cl-3       & 52.11$\pm$12.71 & 92.30$\pm$1.16 & \cellcolor[HTML]{C0C0C0}{\fbseries93.26$\pm$1.02} & 92.78$\pm$1.16 & 92.42$\pm$1.27 & 91.89$\pm$0.79 & 92.98$\pm$1.21 \\
news-5cl-1       & 27.06$\pm$7.41 & 88.54$\pm$1.24 & \cellcolor[HTML]{C0C0C0}{\fbseries88.94$\pm$0.89} & 88.78$\pm$1.02 & 88.57$\pm$1.03 & 88.16$\pm$0.91 & 88.87$\pm$0.96 \\
news-5cl-2       & 30.23$\pm$8.69 & 82.73$\pm$1.47 & \cellcolor[HTML]{C0C0C0}{\fbseries83.94$\pm$1.32} & 83.26$\pm$1.20 & 82.66$\pm$1.64 & 81.66$\pm$1.38 & 83.70$\pm$1.27 \\
news-5cl-3       & 25.76$\pm$7.35 & 82.28$\pm$1.61 & \cellcolor[HTML]{C0C0C0}{\fbseries83.54$\pm$1.32} & 83.17$\pm$1.17 & 82.43$\pm$1.75 & 77.62$\pm$1.46 & 83.52$\pm$1.20 \\
\hline
\end{tabular}}
\end{center}
\caption{\footnotesize{Classification accuracy in percent $\pm$ standard deviation for the various classification methods, obtained on the different datasets. Results are reported for the four feature sets (5\%, 10\%, 20\%, and Ker). For each dataset and method, the final accuracy and standard deviation are obtained by averaging over 10 runs of a standard cross-validation procedure. Each run consists of a nested cross-validation with 5 external folds (test sets, for validation) on which the accuracy and the standard deviation of the classifier are averaged, and 5 internal folds (for parameter tuning). The best performing method is highlighted in boldface for each dataset and each feature set. Bold values highlighted in grey indicate the best performance overall for each dataset, across all feature sets.}}
\label{table:classres}
\end{table}

In order to have a more general overview of the results, a Borda ranking of the methods is performed for each feature set and reported in Table \ref{table:Borda}. The Borda ranking starts by sorting all the methods in ascending order of classification accuracy for each dataset. Then, the score of each method is computed by adding its ranks over all datasets. Therefore, the best method is the one with the highest Borda score reflecting a higher global accuracy across all the datasets.

From Table \ref{table:Borda}, we observe that the ranking of the methods does not change much across the feature sets. The top three methods are always the cBoPH, the FE, and the Sup. For 5\%F and 10\%F, the cBoPH is first followed by the FE and thereafter by the Sup. The cBoPH only exchanges its first rank for the second one with the FE for 20\%F, whereas the Sup takes the second position of the FE for Ker. The fourth and the fifth positions of the ranking are respectively taken by the cBoP and the RSP for 5\%F and Ker, although they exchange their ranks for 20\%F. The two methods on the bottom of the ranking are the SP and the CT. The CT is at the last position except for 5\%F where it is the SP. Globally, the cBoPH is in the first position, followed in order by the FE, the Sup, the cBoP, the RSP, the SP, and the CT. Furthermore, by observing the scores, we notice that the SP and the CT obtain much worse results in comparison to the other methods.

\begin{table}[t!]
\footnotesize
\begin{center}
\scalebox{0.9}{
\begin{tabular}{l|c|c|c|c|c|c|c|c||c|c|}
\cline{2-11}
                              & \multicolumn{2}{c|}{\textbf{5\%}} & \multicolumn{2}{c|}{\textbf{10\%}} & \multicolumn{2}{c|}{\textbf{20\%}} & \multicolumn{2}{c||}{\textbf{Ker}} & \multicolumn{2}{c|}{\textbf{Overall}} \\ \hline
\multicolumn{1}{|l|}{\textbf{Method}} & Score      & Position      & Score       & Position      & Score       & Position      & Score       & Position      & Score       & Position       \\ \hline
\multicolumn{1}{|l|}{cBoPH}   & 83         & 1             & 82          & 1             & 76          & 2             & 84          & 1             & 325         & 1              \\ \hline
\multicolumn{1}{|l|}{FE}      & 73         & 2            & 74          & 2             & 83          & 1             & 67         & 3             & 297         & 2              \\ \hline
\multicolumn{1}{|l|}{Sup}     & 69         & 3             & 63          & 3             & 66          & 3             & 78          & 2             & 276         & 3              \\ \hline
\multicolumn{1}{|l|}{cBoP}    & 59         & 4             & 62          & 4             & 58         & 5             & 57          & 4             & 236         & 4              \\ \hline
\multicolumn{1}{|l|}{RSP}     & 48         & 5             & 60          & 5             & 60       & 4             & 53          & 5             & 221         & 5              \\ \hline
\multicolumn{1}{|l|}{SP}      & 29      & 7             & 28          & 6             & 30          & 6             & 32          & 6             & 119         & 6              \\ \hline
\multicolumn{1}{|l|}{CT}      & 31         & 6             & 23          & 7             & 20          & 7             & 21          & 7             & 95          & 7              \\ \hline
\end{tabular}}
\end{center}
\caption{\footnotesize{Overall position of the different classification techniques for the four feature sets (5\%, 10\%, 20\% and kernel), and overall, according to Borda$'$s method performed across all datasets (the higher the score, the better).}}
\label{table:Borda}
\end{table}

\begin{figure}[t!]
    \centering
    \subfigure[5\%F]{\includegraphics*[width=0.49\textwidth,trim= 0 15 0 0]{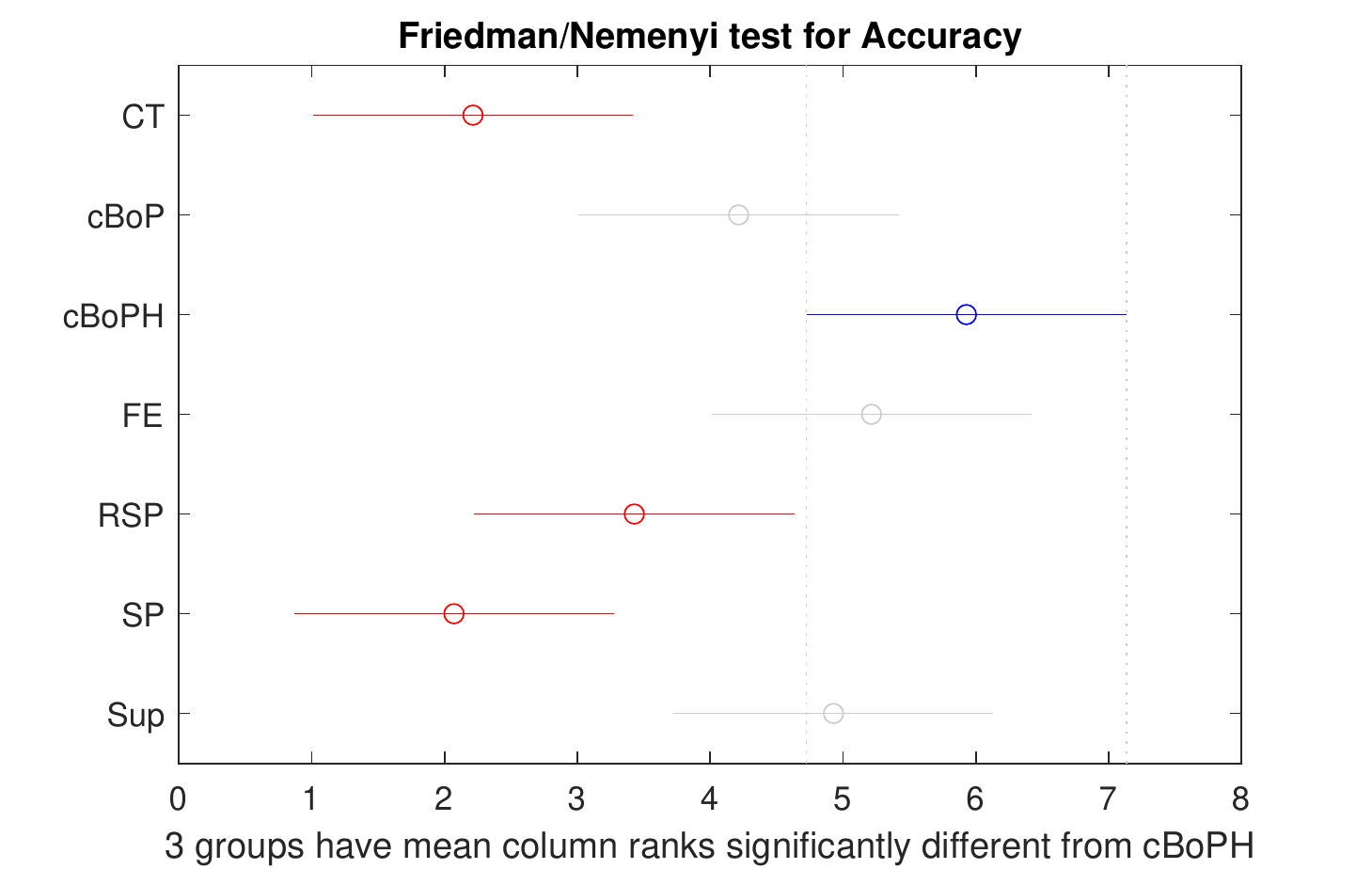}\label{fig:Nemenyi5}}
    \subfigure[10\%F]{\includegraphics*[width=0.49\textwidth,trim= 0 15 0 0]{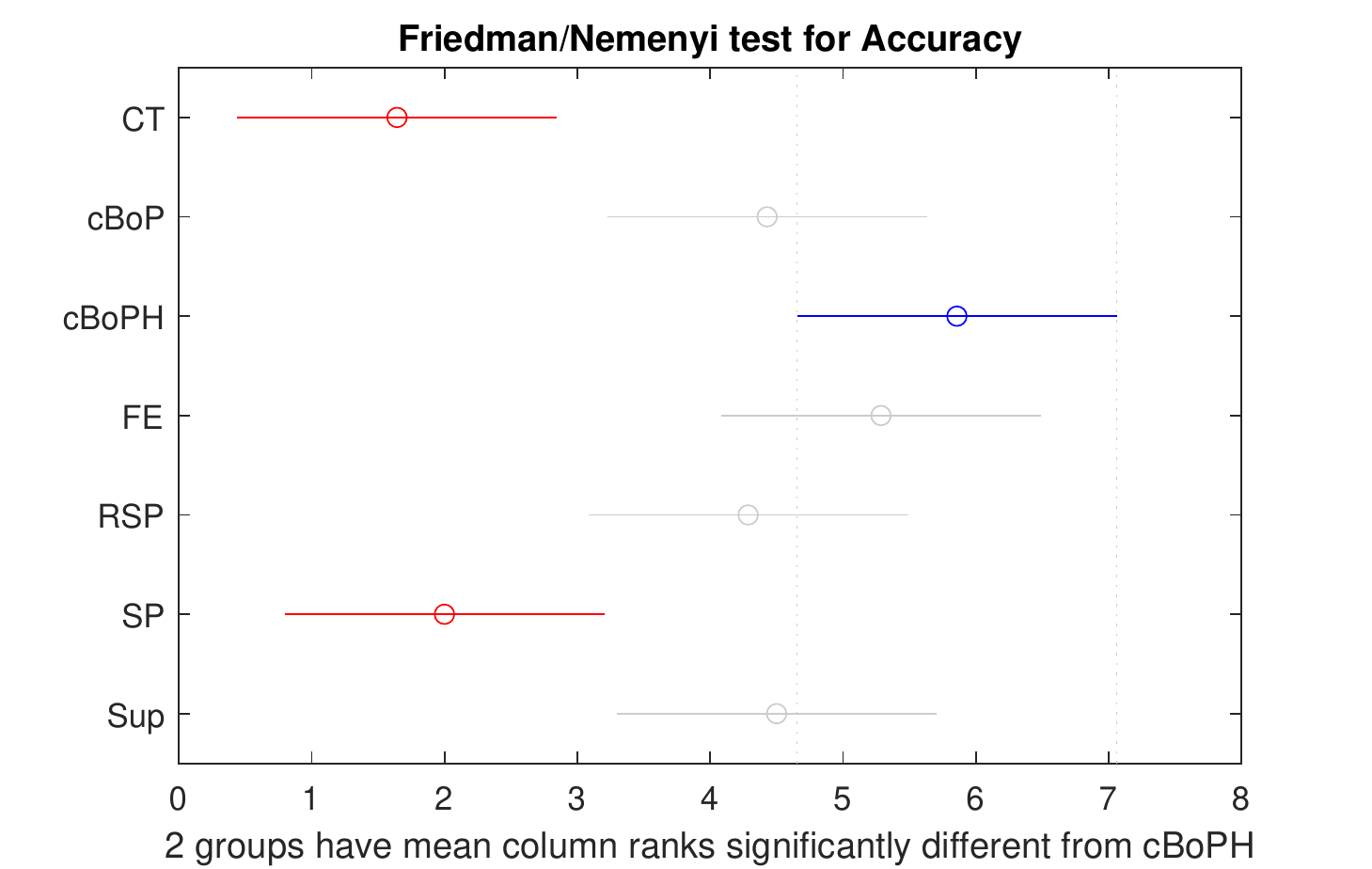}\label{fig:Nemenyi10}}
    \subfigure[20\%F]{\includegraphics*[width=0.49\textwidth,trim= 0 15 0 0]{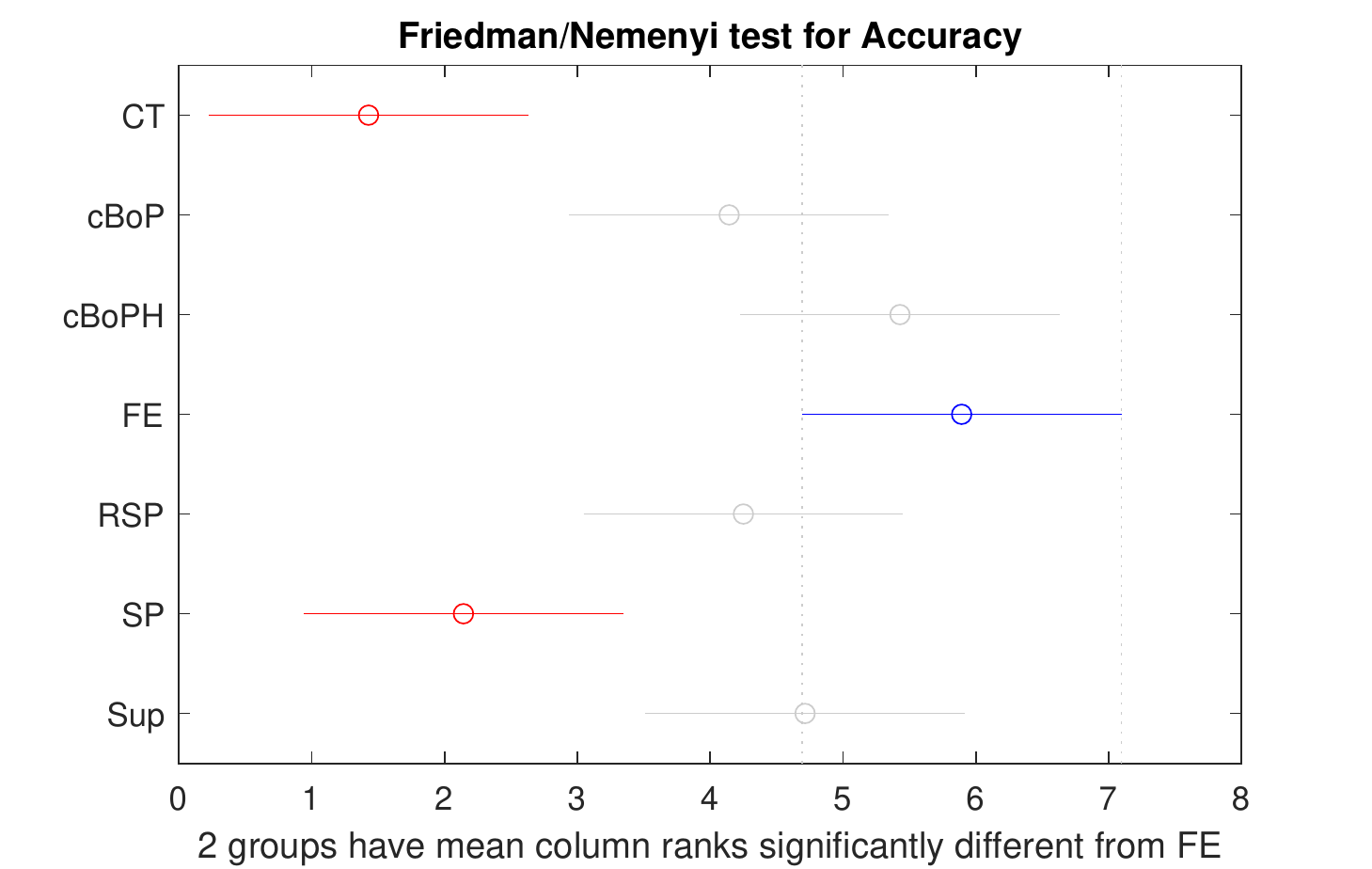}\label{fig:Nemenyi20}}
    \subfigure[Ker]{\includegraphics*[width=0.49\textwidth,trim= 0 15 0 0]{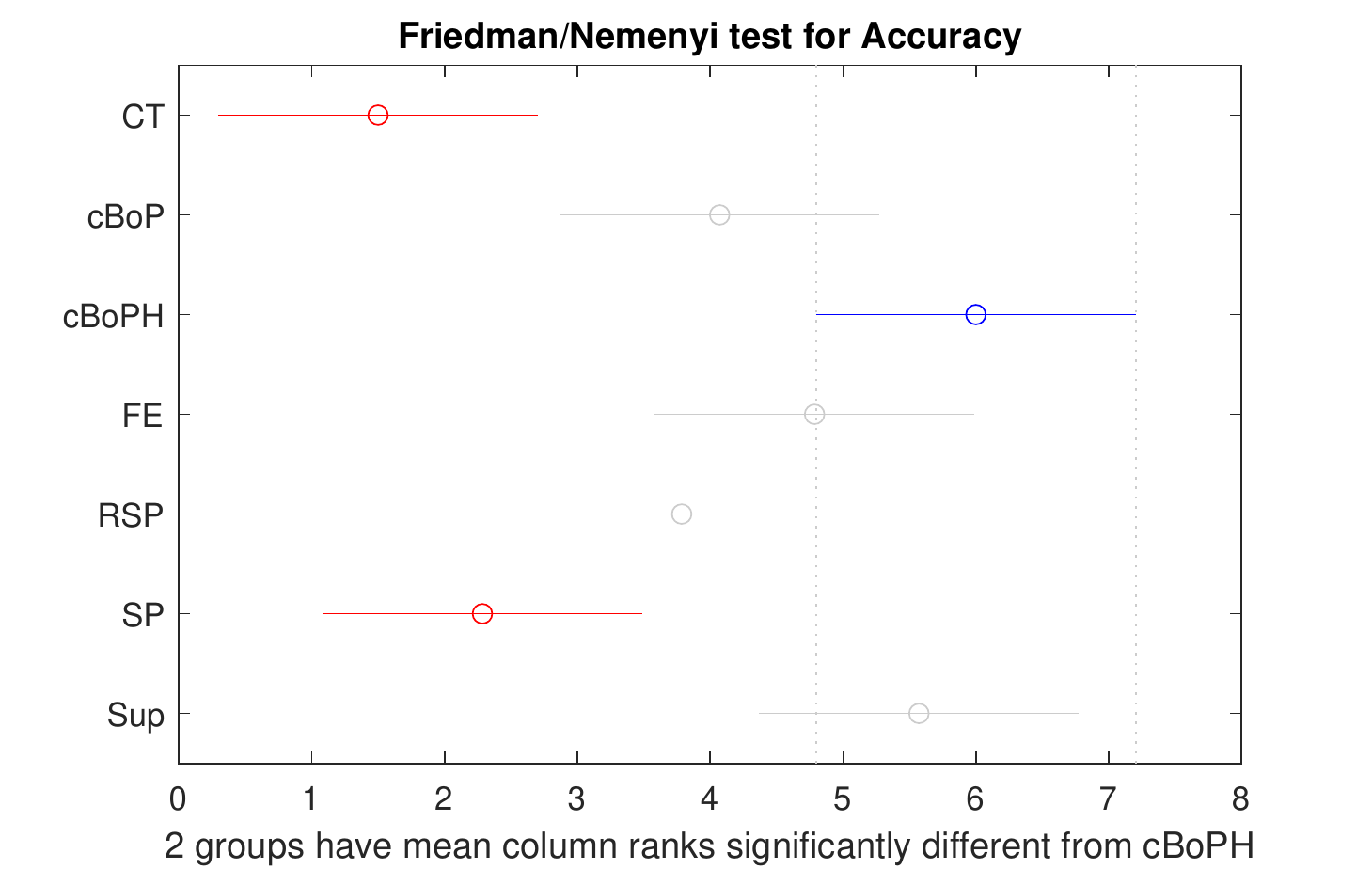}\label{fig:NemenyiKer}}
    \subfigure[Overall]{\includegraphics*[width=0.49\textwidth,trim= 0 15 0 0]{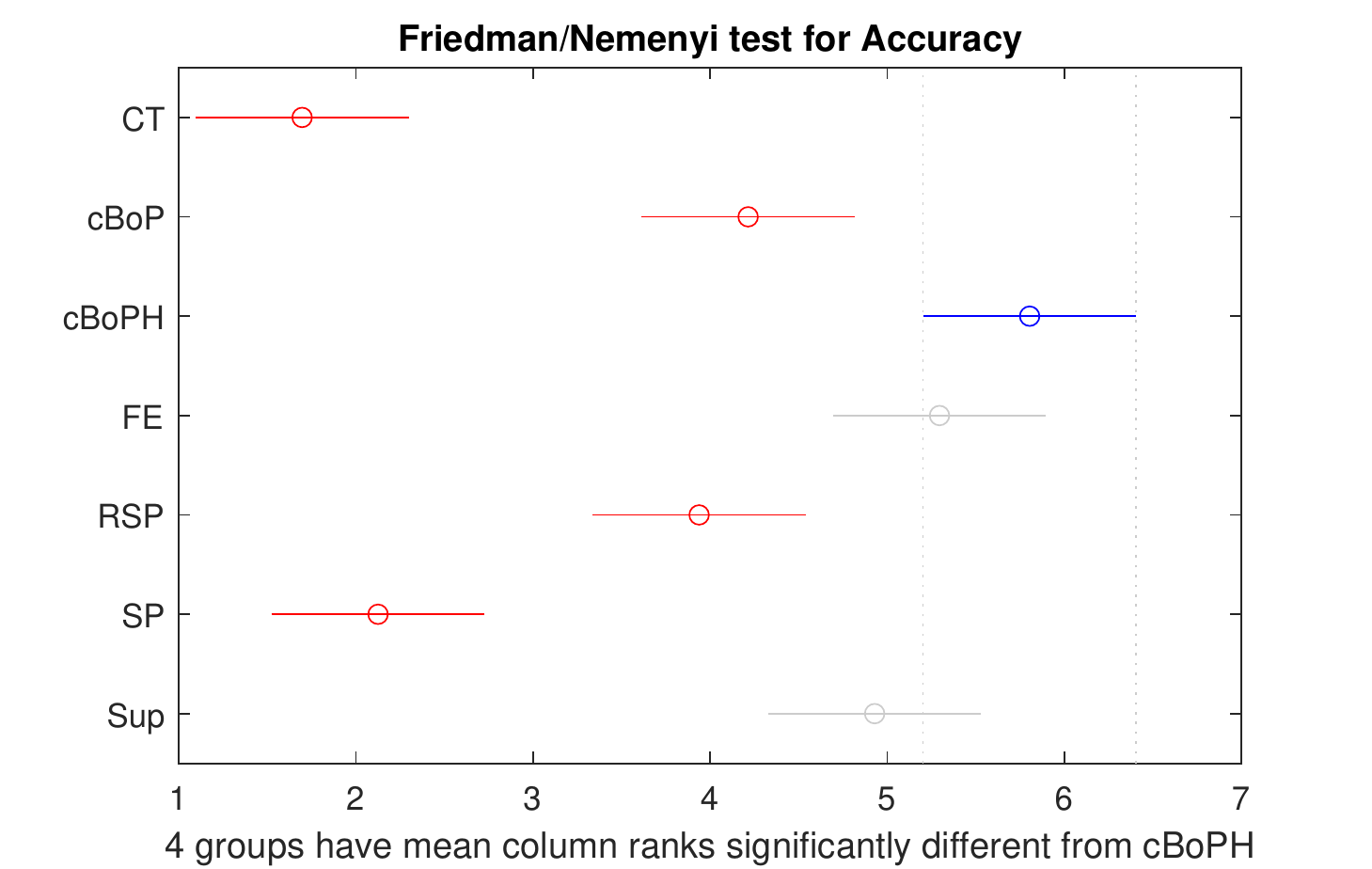}\label{fig:NemenyiOverall}}
    \caption{\footnotesize{Mean ranks and 95\% Nemenyi confidence intervals for the 7 methods across the 14 datasets for feature sets 5\%F (a), 10\%F (b), 20\%F (c), Ker (d) and Overall (all feature sets) (e). Two methods are considered as significantly different if their confidence intervals do not overlap. The axis-x unit is the average rank of the methods. The higher the rank, the best the method. The best method is highlighted.}}
    \label{fig:NemenyiAll}
\end{figure}

The next step of our analysis consists of comparing the different methods across all the 14 datasets through a Friedman test followed by a Nemenyi post-hoc test \cite{Demsar-2006}. The Friedman test is a non-parametric equivalent of the repeated-measures ANOVA. The null hypothesis (H0) of this test is that all the classifiers have the same average ranks. The $p$-values of the Friedman tests are respectively $4.5 \times 10^{-7}$ for 5\%F, $3.7 \times 10^{-8}$ for 10\%F, $7.8 \times 10^{-9}$ for 20\%F, and $7.5 \times 10^{-9}$ for Ker. All these $p$-values are lower than the threshold $\alpha$ of $0.05$, meaning that we can reject H0 and that at least one classifier is significantly different from the others. As all the Friedman tests are positive, we can perform Nemenyi tests, which determine whether or not the performance of each method differs significantly from another. The results of these tests are reported in Figures \ref{fig:Nemenyi5} to \ref{fig:NemenyiKer}. First of all, we can observe that the rankings provided by the Nemenyi tests are quite similar to those provided by the Borda ranking. The tests confirm that the cBoPH, the FE, and the Sup all provide good results, which are significantly superior to the results obtained by the SP and the CT in all feature sets. Moreover, the cBoPH also outperforms the RSP in 5\%F. As regards the cBoP, it performs significantly better than the CT in all feature sets except for 5\%F. Furthermore, we can notice that the cBoP outperforms the SP for 10\%F. The tests also show that the RSP obtains results significantly superior to those of the CT for 10\%F and 20\%F.

We continue our analysis by performing multiple Wilcoxon signed-ranks tests for matched data \cite{Demsar-2006} to potentially discover other significant pairwise differences between the methods. The Wilcoxon signed-ranks test is a non-parametric equivalent of the paired $t$-test. The null hypothesis (H0) of this test is that the two compared classifiers perform equally well. The results of these tests are presented in Table \ref{table:WilcoxonDim5-10} for 5\%F and 10\%F and in Table \ref{table:WilcoxonDim20-Ker} for 20\%F and Ker. All the $p$-values lower than our threshold $\alpha$ of $0.05$ are highlighted in boldface indicating that H0 is rejected. Besides confirming the findings of the Friedman-Nemenyi tests, the Wilcoxon tests show that the SP is outperformed by all the methods except the CT in all feature sets. As regards the CT, it is as well outperformed by all the methods except the SP for 5\%F, 10\%F, and 20\%F. Moreover, the CT obtains results significantly inferior to those of all the methods in Ker. These findings confirm that the techniques developed in the bag-of-paths framework can take advantage of both the SP and the CT to outperform them whatever the retained amount of information. The tests also highlight that the cBoPH performs significantly better than the Sup in 10\%F and 20\%F and the RSP in 10\%F and Ker. On its side, the RSP obtains results significantly inferior to those of the Sup for 5\%F, and Ker as well as those of the FE in all feature sets, except for 10\%F. 
\begin{table}[H]
\footnotesize
\begin{center}
\scalebox{1}{
\begin{tabular}{|l|c|c|c|c|c|c|c|}
\hline
\textbf{Method} & CT                                              & cBoP                                            & cBoPH                          & FE                             & RSP                                             & SP                             & Sup                            \\ \hline
CT              & \cellcolor[HTML]{000000}{\color[HTML]{000000} } & \cellcolor[HTML]{C0C0C0}0.0012                  & \cellcolor[HTML]{C0C0C0}0.0006 & \cellcolor[HTML]{C0C0C0}0.0006 & \cellcolor[HTML]{C0C0C0}0.0006                  & 0.9515                         & \cellcolor[HTML]{C0C0C0}0.0006 \\ \hline
cBoP            & \cellcolor[HTML]{C0C0C0}0.0002                  & \cellcolor[HTML]{000000}{\color[HTML]{000000} } & 0.6257                         & 0.6698                         & 0.4631                                          & \cellcolor[HTML]{C0C0C0}0.0023 & 0.8077                         \\ \hline
cBoPH           & \cellcolor[HTML]{C0C0C0}0.0001                  & 0.5830                                          & \cellcolor[HTML]{000000}       & 0.3910                         & \cellcolor[HTML]{C0C0C0}0.0002                  & \cellcolor[HTML]{C0C0C0}0.0002 & \cellcolor[HTML]{C0C0C0}0.0085 \\ \hline
FE              & \cellcolor[HTML]{C0C0C0}0.0001                  & 0.6257                                          & 0.2166                         & \cellcolor[HTML]{000000}       & \cellcolor[HTML]{C0C0C0}0.0001                  & \cellcolor[HTML]{C0C0C0}0.0006 & 0.3258                         \\ \hline
RSP             & \cellcolor[HTML]{C0C0C0}0.0001                  & 0.5830                                          & \cellcolor[HTML]{C0C0C0}0.0419 & 0.0906                         & \cellcolor[HTML]{000000}{\color[HTML]{000000} } & \cellcolor[HTML]{C0C0C0}0.0006 & \cellcolor[HTML]{C0C0C0}0.0085 \\ \hline
SP              & 0.6257                                          & \cellcolor[HTML]{C0C0C0}0.0012                  & \cellcolor[HTML]{C0C0C0}0.0004 & \cellcolor[HTML]{C0C0C0}0.0004 & \cellcolor[HTML]{C0C0C0}0.0006                  & \cellcolor[HTML]{000000}       & \cellcolor[HTML]{C0C0C0}0.0004 \\ \hline
Sup             & \cellcolor[HTML]{C0C0C0}0.0001                  & 0.8077                                          & \cellcolor[HTML]{C0C0C0}0.0017 & 0.1353                         & 0.7609                                          & \cellcolor[HTML]{C0C0C0}0.0006 & \cellcolor[HTML]{000000}       \\ \hline
\end{tabular}}
\end{center}
\caption{\footnotesize{The $p$-values provided by a pairwise Wilcoxon signed-rank test, for 5\%F in the upper right triangle and the 10\%F in the lower left.}}
\label{table:WilcoxonDim5-10}
\end{table}

\begin{table}[H]
\footnotesize
\begin{center}
\scalebox{1}{
\begin{tabular}{|l|c|c|c|c|c|c|c|}
\hline
\textbf{Method} & CT                                              & cBoP                                            & cBoPH                          & FE                             & RSP                                             & SP                             & Sup                            \\ \hline
CT              & \cellcolor[HTML]{000000}{\color[HTML]{000000} } & \cellcolor[HTML]{C0C0C0}0.0002                  & \cellcolor[HTML]{C0C0C0}0.0001 & \cellcolor[HTML]{C0C0C0}0.0001 & \cellcolor[HTML]{C0C0C0}0.0001                  & \cellcolor[HTML]{FFFFFF}0.5416 & \cellcolor[HTML]{C0C0C0}0.0001 \\ \hline
cBoP            & \cellcolor[HTML]{C0C0C0}0.0006                  & \cellcolor[HTML]{000000}{\color[HTML]{000000} } & 0.5016                         & \cellcolor[HTML]{FFFFFF}0.3258 & \cellcolor[HTML]{FFFFFF}0.5830                  & \cellcolor[HTML]{C0C0C0}0.0012 & \cellcolor[HTML]{FFFFFF}0.9032 \\ \hline
cBoPH           & \cellcolor[HTML]{C0C0C0}0.0001                  & \cellcolor[HTML]{FFFFFF}0.6257                  & \cellcolor[HTML]{000000}       & \cellcolor[HTML]{FFFFFF}0.3258 & \cellcolor[HTML]{FFFFFF}0.2958                  & \cellcolor[HTML]{C0C0C0}0.0004 & \cellcolor[HTML]{C0C0C0}0.0494 \\ \hline
FE              & \cellcolor[HTML]{C0C0C0}0.0004                  & \cellcolor[HTML]{FFFFFF}0.8552                  & \cellcolor[HTML]{FFFFFF}0.0906 & \cellcolor[HTML]{000000}       & \cellcolor[HTML]{C0C0C0}0.0327                  & \cellcolor[HTML]{C0C0C0}0.0004 & \cellcolor[HTML]{FFFFFF}0.1040 \\ \hline
RSP             & \cellcolor[HTML]{C0C0C0}0.0006                  & \cellcolor[HTML]{FFFFFF}0.9032                  & \cellcolor[HTML]{C0C0C0}0.0009 & \cellcolor[HTML]{C0C0C0}0.0295 & \cellcolor[HTML]{000000}{\color[HTML]{000000} } & \cellcolor[HTML]{C0C0C0}0.0006 & \cellcolor[HTML]{FFFFFF}0.8552 \\ \hline
SP              & \cellcolor[HTML]{C0C0C0}0.0006                  & \cellcolor[HTML]{C0C0C0}0.0017                  & \cellcolor[HTML]{C0C0C0}0.0002 & \cellcolor[HTML]{C0C0C0}0.0002 & \cellcolor[HTML]{C0C0C0}0.0006                  & \cellcolor[HTML]{000000}       & \cellcolor[HTML]{C0C0C0}0.0004 \\ \hline
Sup             & \cellcolor[HTML]{C0C0C0}0.0001                  & \cellcolor[HTML]{FFFFFF}0.7148                  & \cellcolor[HTML]{FFFFFF}0.1726 & \cellcolor[HTML]{FFFFFF}0.2166 & \cellcolor[HTML]{C0C0C0}0.0134                  & \cellcolor[HTML]{C0C0C0}0.0002 & \cellcolor[HTML]{000000}       \\ \hline
\end{tabular}}
\end{center}
\caption{\footnotesize{The $p$-values provided by a pairwise Wilcoxon signed-rank test, for 20\%F in the upper right triangle and Ker in the lower left.}}
\label{table:WilcoxonDim20-Ker}
\end{table}
\noindent
Finally, for information, we also analyze the results overall by concatenated the 56 results obtained across the 14 datasets and the four feature sets for each method (last drawing, Figure \ref{fig:NemenyiOverall}). Here, the assumption that the 56 datasets are independent of each other is certainly not fulfilled (they are partially overlapping), so that we cannot draw any statistical conclusion. 
However, we can notice that these results confirm the findings of the Borda ranking.

In summary, the first part of the experiments showed that three methods stand out from the others: the cBoPH, the FE, and the Sup. These methods achieve to consistently outperform most of the methods through all the feature sets on the investigated datasets. Among these three, the cBoPH set itself apart by being the best method across three of the four feature sets and the second in the last one according to the Borda ranking (see Table \ref{table:Borda}).

\subsubsection{Comparison of the impact of the different extracted feature sets}

We now analyze the impact of the feature extraction technique (with a growing number of extracted features) on the classification results, limited to the three methods performing best in the first part of the experiments for conciseness. Nevertheless, we have performed the analysis on all the methods and have drawn similar conclusions except for the CT.  As we already pointed out, the performances of the CT are lower in Ker compared to the other feature sets which is not the case of the other methods. 

\begin{figure}[H]
    \centering
    \subfigure[cBoPH]{\includegraphics*[width=0.49\textwidth,trim= 0 15 0 0]{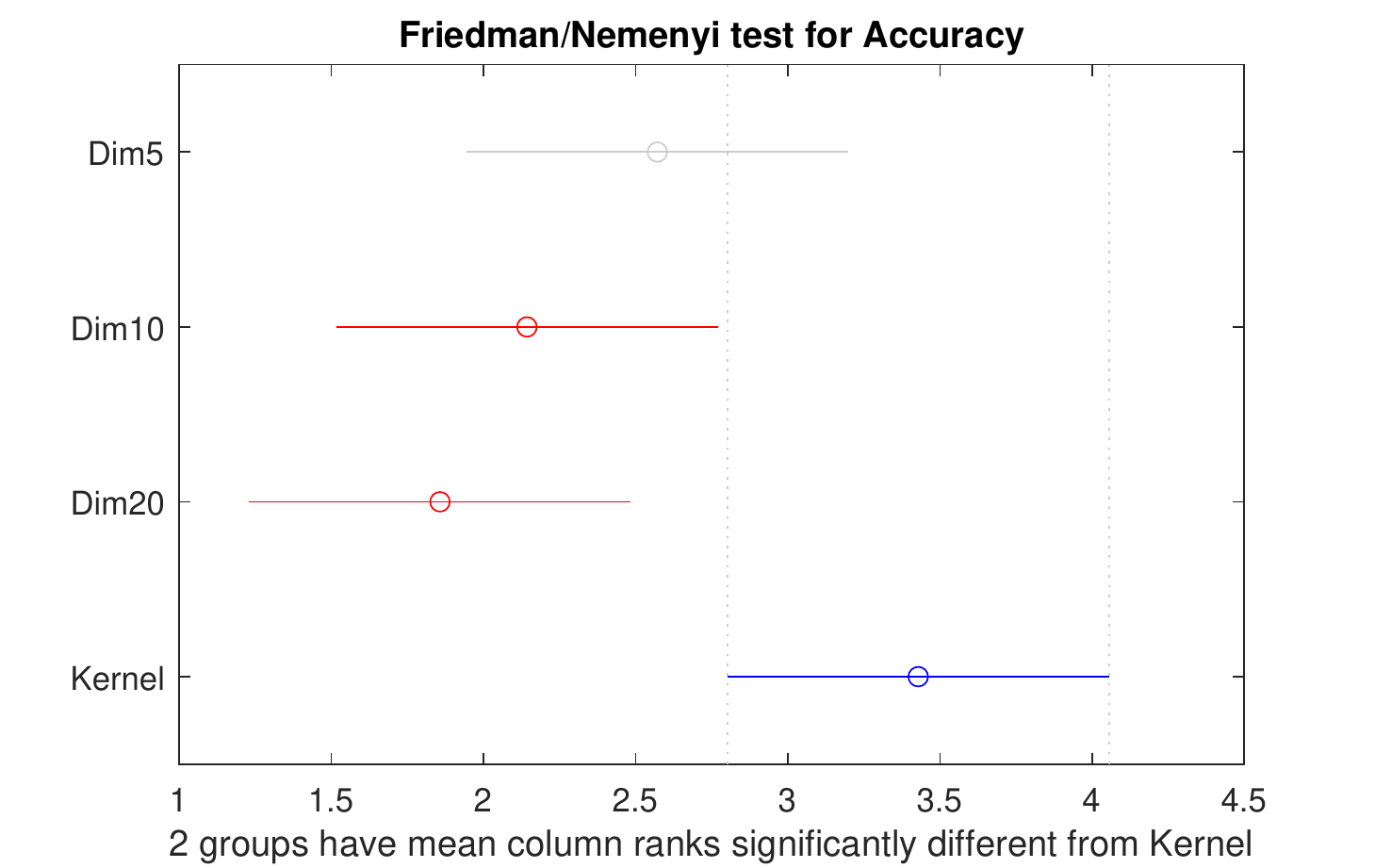}\label{fig:NemenyicBoPH}}
    \subfigure[FE]{\includegraphics*[width=0.49\textwidth,trim= 0 15 0 0]{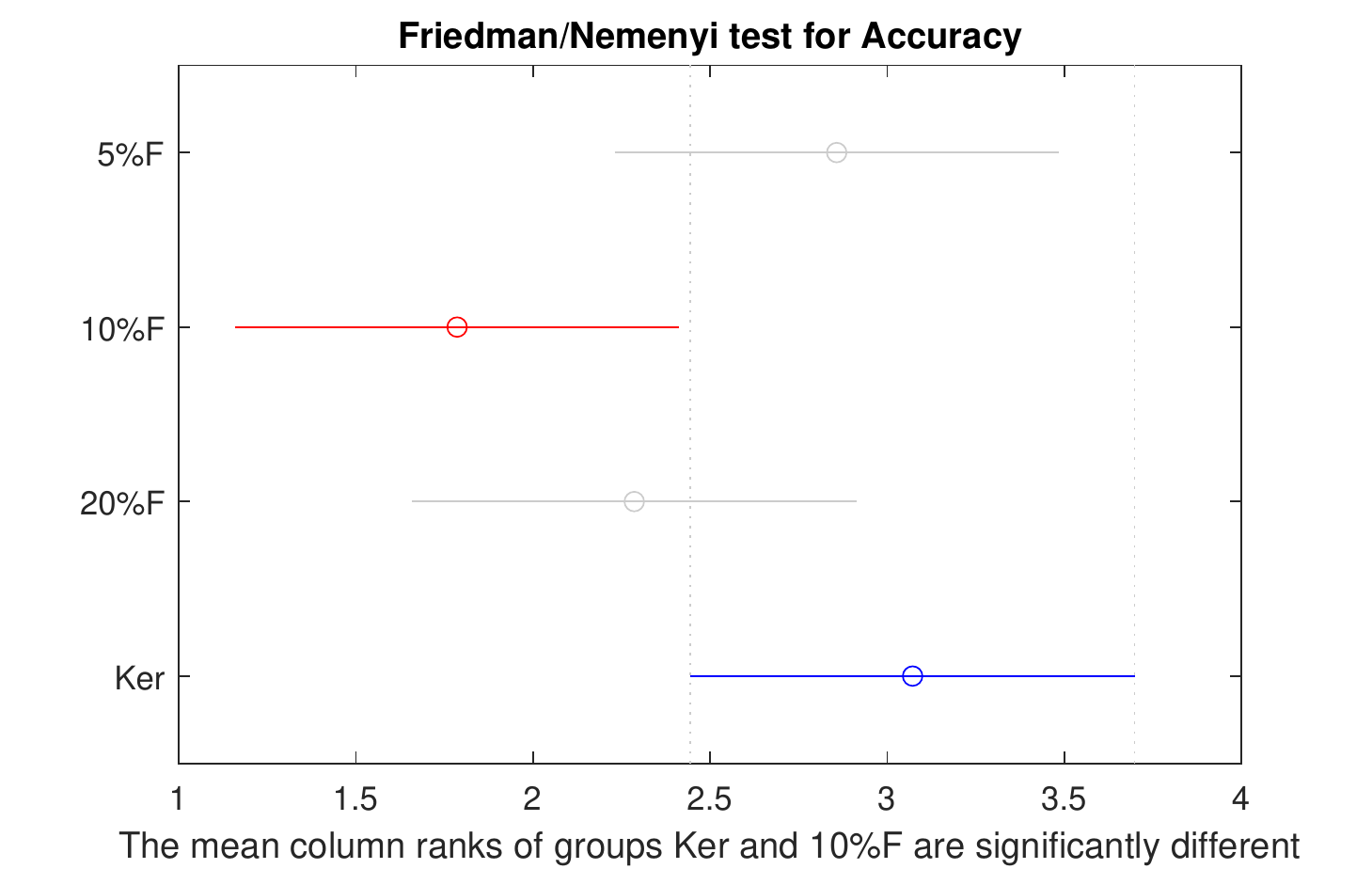}\label{fig:NemenyiFE}}
    \subfigure[Sup]{\includegraphics*[width=0.49\textwidth,trim= 0 15 0 0]{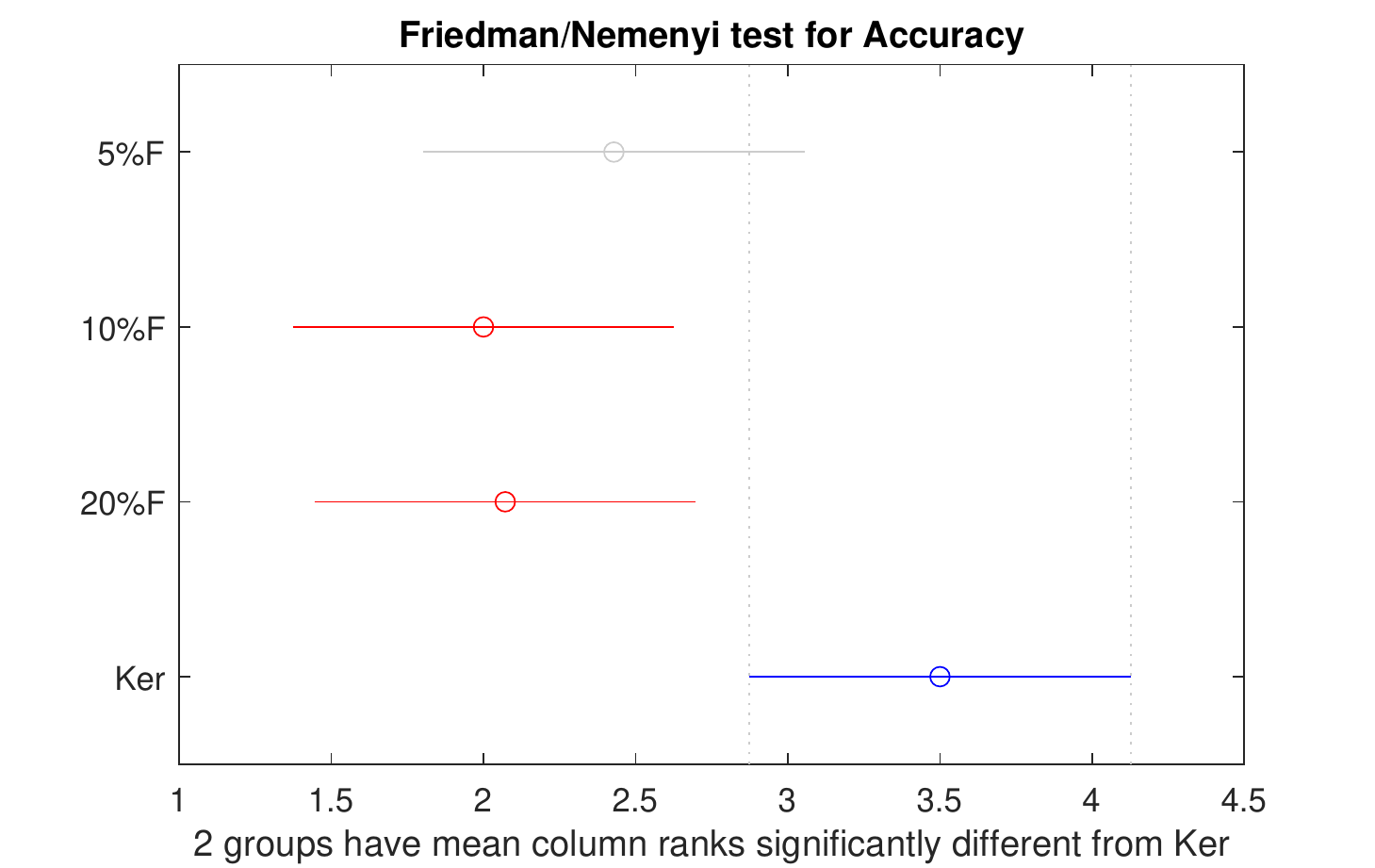}\label{fig:NemenyiSup}}
    \caption{\footnotesize{Mean ranks and 95\% Nemenyi confidence intervals for the three best methods across the four feature sets (5\%F, 10\%F, 20\%F, and Ker). Two feature sets are considered as significantly different if their confidence intervals do not overlap. The axis-x unit is the average rank of the feature sets: the higher the rank, the best the results obtained on the feature set. Each time, the best feature set is highlighted.}}
    \label{fig:NemenyiAllIntra}
\end{figure}

To analyze the results of the three methods across all the feature sets ($5\%$, $10\%$, $20\%$ and Ker), we followed the same procedure as before. First, we perform Friedman tests to identify if there are some differences between the feature sets for each method. As all the $p$-values of these tests are lower than our threshold $\alpha$ of $0.05$\footnote{The $p$-values of the Friedman tests are respectively $0.0080$ for the cBoPH, $0.0370$ for the FE, $0.0169$ and $0.0071$ for the Sup.}, we pursue our analysis by performing Nemenyi tests \cite{Demsar-2006} and reported the results in Figures \ref{fig:NemenyicBoPH} to \ref{fig:NemenyiSup}. From these figures, we can observe that the Ker feature extraction outperforms the 10\%F and the 20\%F feature sets for the Sup and the cBoPH methods. We can also notice that the results obtained by the Ker feature set are significantly superior to those of the 10\%F feature set for the FE method. 

To refine our analysis, we also perform multiple Wilcoxon signed-ranks tests performing pairwise comparisons \cite{Demsar-2006}. The Wilcoxon tests show that the Ker feature set outperforms (again at the $\alpha = 0.05$ level) the 20\%F feature set for the FE ($p$-value = 0.0494) method. Moreover, we can also observe that the 5\%F feature set seems to outperform the 10\%F feature set for the Sup ($p$-value = 0.0494). Nevertheless, the tests do not show any significantly difference between the Ker feature set and the 5\%F feature set for the cBoPH ($p$-value = 0.3910), the FE ($p$-value = 0.8552) and the Sup ($p$-value = 0.2412).

To conclude, the second part of the experiments highlighted that the results obtained by extracting 5\% of dominant eigenvectors are not significantly different from the results obtained by using all the information contained in the kernel for all the methods except the CT. This finding seems to show that the techniques developed in the bag-of-paths framework can perform equally well with only some parts of the information than with all of them, at least on the investigated datasets and following the stated methodology.
\section{Conclusion}
\label{Sec_Conclusion01}

In this work, we introduced a new algorithm that solves the relative entropy-regularized minimum expected cost flow with capacity constraint problem on a graph. This new formulation of the problem extends the previous ones \cite{Guex-2016,Guex-2019} by integrating flow capacity constraints, which frequently appear in real-world applications \cite{Ahuja-1993}.  Therefore, this contribution expands the applications of the previous models to a larger range of real-world applications. 

Furthermore, the first part of the experimental comparisons demonstrated that the margin-constrained bag-of-paths surprisal distance and its hitting version are competitive in comparison with other bag-of-paths methods and, consequently, with other state-of-the-art techniques \cite{Francoisse-2017,Guex-2021,Sommer-2016}.

In addition, the second part of the experiments shows that the performance of the two best bag-of-paths methods does not decrease significantly with partial information about the graph structure, in comparison with all the available information in our semi-supervised classification tasks.

Future work will aim at developing a similar approach for a bag of hitting paths, instead of regular paths in the present work. Another interesting track would be to make the link between the present approach and electrical current in the case of an undirected graph.

\section*{Acknowledgements}

\noindent This work was partially supported by the Immediate and the Brufence projects funded by InnovIris (Brussels region), as well as former projects funded by the Walloon region, Belgium, and the Research Council of Norway. We thank these institutions for giving us the opportunity to conduct both fundamental and applied research. 
\begin{center}
\rule{2.5in}{0.01in}
\end{center}

\appendix
\section*{Appendices}
\label{Appendix}
\numberwithin{equation}{section}

These appendices provide some derivations of relevant quantities introduced in the main text as well as basic quantities related to the standard randomized shortest paths model.

\section{Computing transition probabilities compatible with output flow} 
\label{Sec_computing_transition_matrix_consistent01}

In this section, we derive the form of the natural transition matrix on the extended graph $G_{\mathrm{ext}}$ in such a way that output flow constraints are satisfied (see Equation (\ref{Eq_flow_constraints01})) by using a consistency argument \cite{Guex-2016}.

\subsection{The transition matrix of the extended graph}

Let us deduce the form of the transition matrix $\Pext$ of the natural random walk on the extended graph as well as the value of $\mathbf{w}$ in Equation (\ref{Eq_extendedAdjacencyMatrix01}). Clearly its first row (source node $1$) is $[0 \; \boldsymbol{\sigma}_{\mathrm{in}}^{\text{T}} \; 0]$ and the last row $n$ (target node) is full of zeros (killing, absorbing, node). The row sums corresponding to the other nodes $i \in \mathcal{V}$ (belonging to the original graph $G$) of the new, extended, $n \times n$ matrix 
$\mathbf{A}_{\mathrm{ext}} = (a^{\mathrm{ext}}_{ij})$ (Equation (\ref{Eq_extendedAdjacencyMatrix01}))
are equal to\footnote{Recall that all nodes with label $i \in \mathcal{V}$ keep the same index in $\mathcal{V}_{\mathrm{ext}}$.} $[\mathbf{A}_{\mathrm{ext}}]_{i \bullet} = a_{i \bullet}^{\mathrm{ext}} = a_{i \bullet} + w_{i}$ with $a_{i \bullet} = \sum_{j \in \mathcal{V}} a_{ij}$. Thus, for $i,j \in \mathcal{V}$ (in other words, $i,j \notin \{ 1,n \}$), $p^{\mathrm{ext}}_{ij} = a_{ij}/(a_{i \bullet} + w_{i})$ and, for $j = n$, $p^{\mathrm{ext}}_{in} = w_{i}/(a_{i \bullet} + w_{i})$.

We now introduce the $(n-2) \times 1$ vector $\boldsymbol{\alpha}$ containing the elements corresponding to nodes $i \in \mathcal{V}$ on the last column $n$ of $\Pext$ (see Equation (\ref{Eq_extendedTransitionMatrix01})),
\begin{equation}
\alpha_{i} \triangleq p^{\mathrm{ext}}_{in} = \frac{w_{i}}{a_{i \bullet} + w_{i}} \text{ for } i \in \mathcal{V}
\label{Eq_alpha_value01}
\end{equation}
hence containing elements $i,n$ of the $n \times n$ transition matrix of the extended graph.
This quantity corresponds to the probability of being killed in node $i$ during the next step of the random walk (transiting to the absorbing/cemetery node $n$). Moreover, it is easy to verify that $1-\alpha_{i} = a_{i \bullet}/(a_{i \bullet} + w_{i})$, which is the probability of surviving (continuing the walk). Then, we readily obtain, now for $i,j \in \mathcal{V}$,
\begin{equation}
p^{\mathrm{ext}}_{ij} = \frac{a_{ij}}{a_{i \bullet} + w_{i}}
= \frac{a_{i \bullet}}{(a_{i \bullet} + w_{i})} \times \frac{a_{ij}}{a_{i \bullet}}
= (1-\alpha_{i}) p_{ij}
\end{equation}
where the $p_{ij}$ are the transition probabilities of the random walk on the original graph $G$ (see Equation (\ref{Eq_transition_probabilities_original_graph01})).
Therefore, the \textbf{transition matrix} representing the killed random walk on the extended graph is
\begin{equation}
 \Pext = \kbordermatrix{
               &          1 &  \{ 2, \dots, (n-1) \} = \mathcal{V}                                     &         n  \cr
1              &          0 & \boldsymbol{\sigma}_{\mathrm{in}}^{\text{T}} &         0  \cr
 \{ 2, \dots, (n-1) \} = \mathcal{V}  & \mathbf{0} & (\mathbf{I} - \mathbf{Diag}(\boldsymbol{\alpha})) \mathbf{P}                                           & \boldsymbol{\alpha} \cr
n              &          0 & \phantom{{}^{\text{T}}} \mathbf{0}^{\text{T}}        &         0  \cr
} \nonumber
\end{equation}
where $\mathbf{Diag}(\boldsymbol{\alpha})$ is a diagonal matrix with $\boldsymbol{\alpha}$ on its diagonal and the elements of the vector $\boldsymbol{\alpha}$ still have to be determined -- this problem is solved in the next subsection. Recall that, in this last equation, the transition matrix $\mathbf{P}$ is provided by the natural random walk on the original graph (see Equation (\ref{Eq_transition_probabilities_original_graph01})). Once $\boldsymbol{\alpha}$ computed, the elements of the weight vector $\mathbf{w}$ appearing in the adjacency matrix of Equation (\ref{Eq_extendedAdjacencyMatrix01}) can be obtained from (\ref{Eq_alpha_value01}),
\begin{equation}
    w_{i} = \frac{\alpha_{i}a_{i \bullet}}{1-\alpha_{i}}
\end{equation}

\subsection{Computing the values of $\boldsymbol{\alpha}$}
\label{Subsec_appendix_A2}

In this subsection, the vector $\boldsymbol{\alpha}$ is computed by following a reasoning similar to the one appearing in \cite{Guex-2016,Guex-2019}.
Let us first recall two fundamental Markov chain quantities defined on $G_{\mathrm{ext}}$, the expected number of visits to node $i \in \mathcal{V}$, denoted as $n_{i}$, and the expected number of passages (also called the flow) through edge $(i,j)$, $n_{ij}$. Note that by conservation of flows, we have $n_{j} = \sum_{i \in \mathcal{P}red(j)} n_{ij}$ and $n_{ij} = n_{i} p^{\mathrm{ext}}_{ij}$ for any $j \in \mathcal{V}$.

The elements $\alpha_{i}$ with $i \in \mathcal{V}$ will be set such that the flows in the output edges $(i,n)$, $n_{in}$, \emph{are equal to the prescribed output flows}, $n_{in} = n_{i} p^{\mathrm{ext}}_{in} = n_{i} \alpha_{i} = \sigma_{i}^{\mathrm{out}}$. Thus, $\alpha_{i} = \sigma_{i}^{\mathrm{out}}/n_{i}$ and we therefore have to find the $n_{i}$.

For computing these $n_{i}$, let us start from the following identity, valid for each $j \ne \{ 1,n \}$ (and thus $j \in \mathcal{V}$) on the extended graph,
\begin{align}
n_{j} &= \sum_{i \in \mathcal{P}red(j)} n_{ij}
 = \sum_{i \in \mathcal{P}red(j)} n_{i} p^{\mathrm{ext}}_{ij} \nonumber \\
 &= \sum_{i=1}^{n} n_{i} p^{\mathrm{ext}}_{ij}
 = \sum_{i=2}^{n} n_{i} p^{\mathrm{ext}}_{ij} + \underbracket[0.5pt][5pt]{n_{1}}_{1} \underbracket[0.5pt][3pt]{p^{\mathrm{ext}}_{1j}}_{\sigma_{j}^{\mathrm{in}}} \nonumber \\
 &= \sum_{i=2}^{n} n_{i} p^{\mathrm{ext}}_{ij} + \sigma_{j}^{\mathrm{in}}
 = \sum_{i=2}^{n-1} n_{i} p^{\mathrm{ext}}_{ij} + n_{n} \underbracket[0.5pt][3pt]{p^{\mathrm{ext}}_{nj}}_{0} + \, \sigma_{j}^{\mathrm{in}} \nonumber \\
  &= \sum_{i \in \mathcal{V}} n_{i} p^{\mathrm{ext}}_{ij} + \sigma_{j}^{\mathrm{in}} \quad \text{for } j \in \mathcal{V}
  \label{Eq_expected_visits01}
\end{align}

But we know from Equation (\ref{Eq_extendedTransitionMatrix01}) that $p^{\mathrm{ext}}_{ij} = (1 - \alpha_{i}) p_{ij}$ when $i,j \notin \{ 1,n \}$, which is injected in Equation (\ref{Eq_expected_visits01}). Recalling also that $n_{i} \alpha_{i} = \sigma_{i}^{\mathrm{out}}$, we obtain
\begin{align}
n_{j} &= \sum_{i \in \mathcal{V}} n_{i} p^{\mathrm{ext}}_{ij} + \sigma_{j}^{\mathrm{in}}
 = \sum_{i \in \mathcal{V}} n_{i} (p_{ij} - \alpha_{i} p_{ij}) + \sigma_{j}^{\mathrm{in}} \nonumber \\
&= \sum_{i \in \mathcal{V}} n_{i} p_{ij} - \sum_{i \in \mathcal{V}} \underbracket[0.5pt][3pt]{n_{i} \alpha_{i}}_{\sigma_{i}^{\mathrm{out}}} p_{ij} + \sigma_{j}^{\mathrm{in}} \nonumber \\
&= \sum_{i \in \mathcal{V}} n_{i} p_{ij} - \sum_{i \in \mathcal{V}} p_{ij} \sigma_{i}^{\mathrm{out}} + \sigma_{j}^{\mathrm{in}} \quad \text{for } j \in \mathcal{V}
\end{align}
or, in matrix form, where $\mathbf{n}$ is the $(n-2) \times 1$ column vector containing the $n_{i}$,
\begin{equation}
\mathbf{n} = \mathbf{P}^{\text{T}} \mathbf{n} - \mathbf{P}^{\text{T}} \boldsymbol{\sigma}_{\mathrm{out}} + \boldsymbol{\sigma}_{\mathrm{in}} \nonumber
\end{equation}

We thus have to solve
\begin{equation}
(\mathbf{I} - \mathbf{P}^{\text{T}}) \mathbf{n} = \boldsymbol{\sigma}_{\mathrm{in}} - \mathbf{P}^{\text{T}} \boldsymbol{\sigma}_{\mathrm{out}}
\label{Eq_equation_expected_number_of_visits01}
\end{equation}
and then compute $\boldsymbol{\alpha}$ from
\begin{equation}
\boldsymbol{\alpha} = \boldsymbol{\sigma}_{\mathrm{out}} \div \mathbf{n}
\label{Eq_computing_alpha_matrix_form01}
\end{equation}
where $\div$ denotes the elementwise division.

As $\mathbf{P}^{\text{T}}$ is rank-deficient, the solution to Equation (\ref{Eq_equation_expected_number_of_visits01}) is the sum of a particular solution plus any vector in the null space of $(\mathbf{I} - \mathbf{P}^{\text{T}})$ \cite{Deutsch-1965,Schott-2005,Searle-1982,Serre-2002}. As the null space of $(\mathbf{I} - \mathbf{P}^{\text{T}})$ is spanned by $\boldsymbol{\pi}$, the equilibrium distribution of the Markov chain with transition matrix $\mathbf{P}$, we obtain
\begin{equation}
\mathbf{n} = \big( \mathbf{I} - \mathbf{P}^{\text{T}} \big)^{+} (\boldsymbol{\sigma}_{\mathrm{in}} - \mathbf{P}^{\text{T}} \boldsymbol{\sigma}_{\mathrm{out}}) + \mu \boldsymbol{\pi}
\label{Eq_matrix_equation_expected_number_of_visits01}
\end{equation}
where  $+$ denotes the Moore-Penrose pseudoinverse and $\mu$ is the non-negative ``persistence parameter" \cite{Guex-2016}. To ensure that $\mathbf{n} \geq \boldsymbol{\sigma}_{\mathrm{out}}$ (so that there is a feasible solution), the parameter $\mu$ has to satisfy the following inequality \cite{Guex-2016},
\begin{equation}
    \mu \geq \max_i \bigg\{ \frac{\sigma_{i}^{\mathrm{out}} - \mathbf{e}_i^{\text{T}} \big( \mathbf{I} - \mathbf{P}^{\text{T}} \big)^{+} (\boldsymbol{\sigma}_{\mathrm{in}} - \mathbf{P}^{\text{T}} \boldsymbol{\sigma}_{\mathrm{out}})}{\pi_i} \bigg\}
\end{equation}

Therefore the transition matrix of the extended graph is provided by Equation (\ref{Eq_extendedTransitionMatrix01}) where $\boldsymbol{\alpha}$ is given by the previous Equations (\ref{Eq_computing_alpha_matrix_form01}) and (\ref{Eq_matrix_equation_expected_number_of_visits01}) and is pre-computed.

\section{Computing quantities of interest from the RSP model}
\label{Appendix:RSP}
For the sake of completeness, this appendix introduces some important quantities that can be derived from the standard randomized shortest paths framework, and is largely inspired by \cite{Courtain-2020,Leleux-2021}. These quantities of interest can be computed by taking the partial derivative of the optimal free energy (see \cite{Fouss-2016,Francoisse-2017,Kivimaki-2012,Saerens-2008,Yen-08K} for details).

\paragraph{Flow in edges.}

For the \textbf{expected number of passages} through edge $(i,j)$, that is, the flow in $(i,j)$ at temperature $T = 1/\theta$, we get from Equations (\ref{Eq_Boltzmann_probability_distribution01}) and (\ref{Eq_optimal_free_energy01}),
\begin{align}
\frac{\partial \phi(\text{P}^{*})}{\partial c_{ij}} &= - \dfrac{1}{\theta \mathcal{Z}} \frac{\partial \mathcal{Z}} {\partial c_{ij}}
 =  - \dfrac{1}{\theta \mathcal{Z}} \sum_{\wp\in\mathcal{P}_{1n}} \tilde{\pi} (\wp)\exp[-\theta \tilde{c}(\wp)] (- \theta) \frac{\partial \tilde{c}(\wp)} {\partial c_{ij}} \nonumber \\
&= \sum_{\wp\in\mathcal{P}_{1n}} \frac{ \tilde{\pi}(\wp) \exp[-\theta \tilde{c}(\wp)] } {\mathcal{Z}} \, \frac{\partial \tilde{c}(\wp)} {\partial c_{ij}} \nonumber \\
&= \sum_{\wp\in\mathcal{P}_{1n}} \text{P}^{*}(\wp) \, \eta\big((i,j) \in \wp\big)
= \bar{n}_{ij}
\end{align}
where the relation $\partial \tilde{c}(\wp)/\partial c_{ij} = \eta\big((i,j) \in \wp\big)$ is used, $\eta\big((i,j) \in \wp\big)$ being the total number of times edge $(i,j)$ appears on path $\wp$. From this result, we can deduce that we obtain, for the flow in $(i,j)$ at temperature $T$,
\begin{equation}
\bar{n}_{ij} = -T \frac{\partial \log\mathcal{Z}} {\partial c_{ij}}
\label{Eq_flows_from_partition_function01}
\end{equation}
For details, see for instance \cite{Francoisse-2017,Kivimaki-2012,Saerens-2008}, but some additional partial results follow in this section.

\paragraph{Fundamental matrix.}

It can be shown that the partition function $\mathcal{Z}$ can be computed in closed form (again, see, e.g., \cite{Francoisse-2017,Kivimaki-2012,Saerens-2008} for details). Let us introduce the \textbf{fundamental matrix} of the randomized shortest paths system,
\begin{equation}
\mathbf{Z} = \mathbf{I} + \mathbf{W} + \mathbf{W}^{2} + \cdots = (\mathbf{I} - \mathbf{W})^{-1}, \quad \text{with } \mathbf{W} = \mathbf{P} \circ \exp[-\theta \mathbf{C}]
\label{Eq_fundamentalMatrix01}
\end{equation}
where we recall that $\mathbf{C}$ is the cost matrix and $\circ$ is the elementwise (Hadamard) product. Elementwise, element $i,j$ of $\mathbf{W}$ is $w_{ij} = p_{ij} \exp[-\theta c_{ij}]$. This expression sums up contributions of different paths lengths, starting from zero-length paths ($\mathbf{I}$). Therefore, $z_{11} = 1$ because node $1$ has no predecessor on the extended graph \cite{Francoisse-2017}.

Moreover, it can also be shown that Equation (\ref{Eq_Boltzmann_probability_distribution01}) can be rewritten in terms of the elements of $\mathbf{W}$ as
\begin{equation}
\text{P}^{*}(\wp) 
= \frac{\tilde{w}(\wp)}{\dsum_{\wp'\in\mathcal{P}_{1n}} \tilde{w}(\wp')}
\label{Eq_Boltzmann_probability_distribution02}
\end{equation}
with $\tilde{w}(\wp)  = \prod_{\tau = 1}^{\ell} w_{\wp(\tau-1) \wp(\tau)} = \prod_{\tau = 1}^{\ell} p_{\wp(\tau-1) \wp(\tau)} \exp[- \theta p_{\wp(\tau-1) \wp(\tau)}] = \tilde{\pi}(\wp) \exp[-\theta \tilde{c}(\wp)]$, that is, the product of the corresponding elements of the matrix along the edges of the path.


\paragraph{Partition function and forward/backward variables.}

In addition, the \textbf{partition function} is simply \cite{Francoisse-2017,Kivimaki-2012,Saerens-2008,Yen-08K}
\begin{equation}
\mathcal{Z} = \dsum_{\wp \in \mathcal{P}_{1n}} \tilde{\pi}(\wp) \exp[-\theta \tilde{c}(\wp)]
= \dsum_{\wp \in\mathcal{P}_{1n}} \tilde{w}(\wp) = [\mathbf{Z}]_{1n} = z_{1n}
\label{Eq_partition_function_definition01}
\end{equation}

More generally \cite{Garcia-Diez-2011}, the following forward and backward variables can be defined,
\begin{equation}
z_{1i} = \dsum_{\wp \in \mathcal{P}_{1i}} \tilde{w}(\wp)
 \quad \text{and}
\quad z_{jn} = \dsum_{\wp \in \mathcal{P}_{jn}} \tilde{w}(\wp)
\label{Eq_forward_backward_variables01}
\end{equation}
where $\mathcal{P}_{1i}$ is the set of paths starting in node $1$ and ending in node $i$ and $\mathcal{P}_{jn}$ is the set of paths starting in node $j$ and ending in node $n$.
Interestingly, the backward variables can be interpreted as probabilities of surviving during a killed random walk with transition matrix $\mathbf{W}$, that is, reaching hitting node $n$ without being killed during the walk \cite{Francoisse-2017}.

\paragraph{Flows and number of visits.}

From (\ref{Eq_flows_from_partition_function01}) and (\ref{Eq_partition_function_definition01}), the directed \textbf{flow} in $(i,j)$ can be obtained from (\ref{Eq_fundamentalMatrix01}) by computing the partial derivative of the logarithm of the partition function with respect to the edge costs,
\begin{equation}
\bar{n}_{ij} = - \tfrac{1}{\theta} \frac{\partial \log\mathcal{Z}} {\partial c_{ij}}
= \frac{ z_{1i} p_{ij}^{\mathrm{ext}} \exp[-\theta c_{ij}]  z_{jn} } {z_{1n}}
= \frac{ z_{1i} w_{ij} z_{jn} } {z_{1n}}
\label{Eq_computation_edge_flows01}
\end{equation}
and since only the first row and the last column of $\mathbf{Z}$ are needed, two systems of linear equations can be solved instead of matrix inversion in Equation (\ref{Eq_fundamentalMatrix01}).
From this last equation, the \textbf{expected number of visits} to a node $j$ can be computed\footnote{Recall that node $1$ has no predecessor and node $n$ no successor.} from
\begin{equation}
\bar{n}_{j} = \sum_{i=1}^{n} \bar{n}_{ij} = \frac{ z_{1j} z_{jn} } {z_{1n}}
\label{Eq_computation_node_flows01}
\end{equation}

\paragraph{Expected cost.}

The \textbf{expected cost} until absorption can also be computed in closed form. By defining the matrix containing the expected number of passages through the edges by $\mathbf{N} = (\bar{n}_{ij})$, the expected cost until reaching the target \cite{Garcia-Diez-2011} is
\begin{equation}
\langle \tilde{c} \rangle = \mathbf{e}^{\text{T}} (\mathbf{N} \circ \mathbf{C}) \mathbf{e}
\label{Eq_real_expected_cost01}
\end{equation}
which corresponds to the sum of the expected number of passages through each edge times the cost of following the edge. Recall that $\circ$ is the elementwise matrix product.

\paragraph{Optimal transition probabilities.}

Finally, the optimal transition probability of following the edge $(i,j)$, minimizing the objective function (\ref{Eq_optimization_problem_BoP01}), is
\begin{equation}
p^{*}_{ij} = \frac{\bar{n}_{ij}}{\bar{n}_{i}} = \frac{z_{jn}}{z_{in}} p_{ij} \exp[-\theta c_{ij}]
\label{Eq_biased_transition_probabilities01}
\end{equation}
It defines a biased random walk on the graph -- the random walker is ``attracted" by the target node $n$. These transition probabilities do not depend on the source node and correspond to the optimal randomized strategy, or policy, minimizing free energy for the given temperature $T$. Equation (\ref{Eq_biased_transition_probabilities01}) corresponds to the counterpart of (\ref{Eq_Boltzmann_probability_distribution01}) at the node level, instead of the path level.

\section{Derivation of the algorithm computing the optimal randomized policy} 
\label{Sec_derivation_of_the_algorithm01}

This appendix section derives the algorithm allowing to compute the randomized policy for the relative entropy-regularized optimal transport on the extended graph $G_{\mathrm{ext}}$, and is directly inspired by \cite{Courtain-2020}. The first part introduces the augmented costs whereas the second part details the iterative procedure allowing to solve the problem.

\subsection{The Lagrange function}
\label{Subsec_Lagrange_function_edge_contraits_details01}

Given that $\bar{n}_{ij} = \sum_{\wp\in\mathcal{P}_{1n}} \text{P}(\wp) \, \eta\big((i,j) \in \wp\big)$ with $\eta\big((i,j) \in \wp\big)$ being the number of times edge $(i,j)$ appears on path $\wp$, the Lagrange function (\ref{Eq_Lagrange_edge_flow_constraints01}) defined on the extended graph $G_{\mathrm{ext}}$ becomes 
\begin{align}
\mathscr{L}(\text{P},\boldsymbol{\lambda})
&= \underbracket[0.5pt][3pt]{ \dsum_{\wp \in \mathcal{P}_{1n}} \text{P}(\wp) \tilde{c}(\wp) + T \dsum_{\wp \in \mathcal{P}_{1n}} \text{P}(\wp) \log \left( \frac{\text{P}(\wp)}{\tilde{\pi}(\wp)} \right) }_{\text{free energy, }\phi(\text{P})}
+ \mu \bigg( \dsum_{\wp \in \mathcal{P}_{1n}} \text{P}(\wp) - 1 \bigg) \nonumber \\
&\quad + \dsum_{i \in \mathcal{I}n} \lambda_{i}^{\mathrm{in}} \bigg[ \dsum_{\wp\in\mathcal{P}_{1n}} \text{P}(\wp) \, \eta\big((1,i) \in \wp\big) - \sigma_{i}^{\mathrm{in}} \bigg] \nonumber \\
&\quad + \dsum_{j \in \mathcal{O}ut} \lambda_{j}^{\mathrm{out}} \bigg[ \dsum_{\wp\in\mathcal{P}_{1n}} \text{P}(\wp) \, \eta\big((j,n) \in \wp\big) - \sigma_{j}^{\mathrm{out}} \bigg]
\label{Eq_Lagrange_edge_flow_constraints02}
\end{align}

Note that the objective function to be minimized is convex and the equality constraints are all linear.
The Lagrange function can be rewritten as
\begin{align}
\mathscr{L}(\text{P},\boldsymbol{\lambda})
&= \dsum_{\wp \in \mathcal{P}_{1n}} \text{P}(\wp) \bigg[ \underbracket[0.5pt][3pt]{ \tilde{c}(\wp) + \dsum_{i \in \mathcal{I}n} \lambda_{i}^{\mathrm{in}} \, \eta\big((1,i) \in \wp\big) + \dsum_{j \in \mathcal{O}ut} \lambda_{j}^{\mathrm{out}} \, \eta\big((j,n) \in \wp\big)  }_{\tilde{c}'(\wp)} \bigg] \nonumber \\
&\quad + T \dsum_{\wp \in \mathcal{P}_{1n}} \text{P}(\wp) \log \left( \frac{\text{P}(\wp)}{\tilde{\pi}(\wp)} \right)
+ \mu \bigg( \dsum_{\wp \in \mathcal{P}_{1n}} \text{P}(\wp) - 1 \bigg) \nonumber \\
&\quad - \dsum_{i \in \mathcal{I}n} \lambda_{i}^{\mathrm{in}} \sigma_{i}^{\mathrm{in}}
- \dsum_{j \in \mathcal{O}ut} \lambda_{j}^{\mathrm{out}} \sigma_{j}^{\mathrm{out}} \nonumber \\
&= \underbracket[0.5pt][3pt]{ \dsum_{\wp \in \mathcal{P}_{1n}} \text{P}(\wp) \tilde{c}'(\wp) + T \dsum_{\wp \in \mathcal{P}_{1n}} \text{P}(\wp) \log \left( \frac{\text{P}(\wp)}{\tilde{\pi}(\wp)} \right) }_{\text{free energy } \phi'(\text{P})}
+ \mu \bigg( \dsum_{\wp \in \mathcal{P}_{1n}} \text{P}(\wp) - 1 \bigg) \nonumber \\
&\quad \underbracket[0.5pt][3pt]{ - \dsum_{i \in \mathcal{I}n} \lambda_{i}^{\mathrm{in}} \sigma_{i}^{\mathrm{in}}
- \dsum_{j \in \mathcal{O}ut} \lambda_{j}^{\mathrm{out}} \sigma_{j}^{\mathrm{out}} }_{\text{does not depend on $\mathrm{P}(\cdot)$}}
\label{Eq_lagrange_function_modified01}
\end{align}
where $\tilde{c}'(\wp) = \sum_{\tau = 1}^{t} c'_{s(\tau-1) s(\tau)}$ is the total cumulated augmented cost along path $\wp$ with the original costs $c'_{ij}$ being replaced by the \emph{augmented costs} on $G_{\mathrm{ext}}$,
\begin{equation}
c'_{ij} =
\begin{cases}
c_{ij}^{\mathrm{ext}} + \lambda_{j}^{\mathrm{in}} \\
c_{ij}^{\mathrm{ext}} + \lambda_{i}^{\mathrm{out}} \\
c_{ij}^{\mathrm{ext}}
\end{cases}
= \begin{cases}
\lambda_{j}^{\mathrm{in}} & \text{when } i = 1 \text{ and } j \in \mathcal{I}n \\
\lambda_{i}^{\mathrm{out}} & \text{when } i \in \mathcal{O}ut \text{ and } j = n \\
c_{ij}^{\mathrm{ext}} & \text{otherwise}
\end{cases} \nonumber
\end{equation}
because the initial costs are equal to zero for edges starting in node $1$ and ending in $\mathcal{I}n$. The same holds for edges starting in $\mathcal{O}ut$ and ending in node $n$. $\mathbf{C}'$ will be the matrix containing these augmented costs.
Besides, $\phi'(\text{P})$ will be the free energy depending on these augmented costs.
This justifies Equation (\ref{Eq_redefined_costs01}).

Thus, in Equation (\ref{Eq_lagrange_function_modified01}), everything happens as if the costs have been redefined by taking into account the Lagrange parameters. These Lagrange parameters can therefore be interpreted as additional costs necessary to satisfy the equality constraints.
We now have to find the Lagrange parameters $\boldsymbol{\lambda}$ by using Lagrangian duality.

\subsection{Exploiting Lagrangian duality}
\label{Subsec_Lagrangian_maximization_dual_01}

From Equation (\ref{Eq_primal_dual_lagrangian01}), we have to compute the dual function and then maximize the dual function in terms of the Lagrange parameters.

\subsubsection{Computing the dual function}

Let us compute $\text{P}^{*} = \argmin_{\{ \text{P}(\wp) \}_{\wp \in \mathcal{P}_{1n}}} \mathscr{L}(\text{P},\boldsymbol{\lambda})$ satisfying the sum-to-one constraint.
From Equation (\ref{Eq_lagrange_function_modified01}), as for the RSP (see Subsection \ref{Sec_randomized_shortest_paths01}), the primal can be solved easily and exactly; it provides a Gibbs-Boltzmann distribution (Equation (\ref{Eq_Boltzmann_probability_distribution01})), but this time in terms of the \emph{augmented costs}, $\tilde{c}'(\wp)$, introduced in Equation (\ref{Eq_redefined_costs01}). It simply computes the new probability distribution $\text{P}^{*}(\cdot)$.

Once $\text{P}^{*}$ has been computed, as shown in Subsection \ref{Sec_randomized_shortest_paths01}, Equation (\ref{Eq_optimal_free_energy01}), $\phi(\text{P}^{*}) = -T \log \mathcal{Z'}$ on $G_{\mathrm{ext}}$, and the corresponding dual function (\ref{Eq_lagrange_function_modified01}) can be rewritten in terms of the partition function (Equation (\ref{Eq_partition_function_definition01})) on $G_{\mathrm{ext}}$ as
\begin{align}
\mathscr{L}(\text{P}^{*},\boldsymbol{\lambda}) &= \phi'(\text{P}^{*}) - \dsum_{i \in \mathcal{I}n} \lambda_{i}^{\mathrm{in}} \sigma_{i}^{\mathrm{in}}
- \dsum_{j \in \mathcal{O}ut} \lambda_{j}^{\mathrm{out}} \sigma_{j}^{\mathrm{out}} \nonumber \\
&= -T \log \mathcal{Z}' - \dsum_{i \in \mathcal{I}n} \lambda_{i}^{\mathrm{in}} \sigma_{i}^{\mathrm{in}}
- \dsum_{j \in \mathcal{O}ut} \lambda_{j}^{\mathrm{out}} \sigma_{j}^{\mathrm{out}} \nonumber
\end{align}
which proves Equation (\ref{Eq_dual_lagrangian01}).

\subsubsection{Maximizing the dual function}

For computing the maximum of the dual function\footnote{Recall that the dual function is concave; see for instance \cite{Griva-2008}.}, we will use a block coordinate ascend \cite{Bertsekas-1999,Lange-2013} procedure optimizing sequentially with respect to the $\{ \lambda_{k}^{\mathrm{in}} \}_{k \in \mathcal{I}n}$ and then with respect to the the $\{ \lambda_{k}^{\mathrm{out}} \}_{k \in \mathcal{O}ut}$. As the dual function is concave, the block coordinate ascend procedure converges to the global maximum.

\paragraph{Computing Lagrange parameters associated to input nodes.}

Let us start with the $\{ \lambda_{k}^{\mathrm{in}} \}_{k \in \mathcal{I}n}$. Recalling that $\partial_{c_{ij}} (-T \log \mathcal{Z}) = \bar{n}_{ij}$ (see Equation (\ref{Eq_computation_edge_flows01})), the gradient of the dual function with respect to $\lambda_{k}^{\mathrm{in}}$ can be computed for each node $k \in \mathcal{I}n$,
\begin{align}
\frac{\partial \mathscr{L}(\text{P}^{*},\boldsymbol{\lambda})}{\partial \lambda_{k}^{\mathrm{in}}} 
&= \frac{\partial (-T \log \mathcal{Z}')} {\partial \lambda_{k}^{\mathrm{in}}} - \sigma_{k}^{\mathrm{in}}
= \frac{\partial (-T \log \mathcal{Z}')} {\partial c'_{1k}} \frac{\partial c'_{1k}} {\partial \lambda_{k}^{\mathrm{in}}} - \sigma_{k}^{\mathrm{in}} \nonumber \\
&= \bar{n}_{1k} - \sigma_{k}^{\mathrm{in}}
\label{Eq_dual_function_derivative_edge_constraints01}
\end{align}
where we used the definition of the augmented costs in Equation (\ref{Eq_redefined_costs01}). By expressing the fact that the gradient must cancel at the optimum simply reduces to the equality constraint of Equation (\ref{Eq_equality_constraints01}), which is common in maximum-entropy problems \cite{Kapur-1989,Kapur-1992} and was already observed in \cite{Courtain-2020},
\begin{equation}
\bar{n}_{1k} = \sigma_{k}^{\mathrm{in}} \quad \text{for each } k \in \mathcal{I}n
\label{Eq_capacity_equality_constraints01}
\end{equation}

Now, from Equation (\ref{Eq_computation_edge_flows01}), $\bar{n}_{1k} = z'_{11} p_{1k}^{\mathrm{ext}} \exp[-\theta c'_{1k}]  z'_{kn} / z'_{1n}$ and, as node $1$ has no incoming edge, $z'_{11} = 1$ (see Equation (\ref{Eq_fundamentalMatrix01})). Moreover, again from Equation (\ref{Eq_fundamentalMatrix01}), $(\mathbf{I} - \mathbf{W}') \mathbf{Z}' = \mathbf{Z}' - \mathbf{W}' \mathbf{Z}' = \mathbf{I}$, and taking element $1$, $n$ provides
\begin{equation}
z'_{1n} = \dsum_{k \in \mathcal{S}ucc(1)} w'_{1k}  z'_{kn}
= \dsum_{k \in \mathcal{S}ucc(1)} p_{1k}^{\mathrm{ext}} \exp[-\theta c'_{1k}]  z'_{kn}
= \dsum_{k \in \mathcal{I}n} p_{1k}^{\mathrm{ext}} \exp[-\theta \lambda_{k}^{\mathrm{in}}]  z'_{kn}
\end{equation}
where $\mathcal{S}ucc(1)$ is the set of successor nodes of node $1$, which corresponds to the set of input nodes $\mathcal{I}n$.
Thus, from the previous discussion and Equation (\ref{Eq_computation_edge_flows01}),
\begin{equation}
\bar{n}_{1k} = \frac{ z'_{11} p_{1k}^{\mathrm{ext}} \exp[-\theta c'_{1k}]  z'_{kn} } {z'_{1n}}
= \frac{ p_{1k}^{\mathrm{ext}} \exp[-\theta \lambda_{k}^{\mathrm{in}}]  z'_{kn} } {\sum_{k' \in \mathcal{I}n} p_{1k'}^{\mathrm{ext}} \exp[-\theta \lambda_{k'}^{\mathrm{in}}]  z'_{k'n}}
\label{Eq_number_passages_in01}
\end{equation}
because $z'_{11} = 1$. We observe that $\sum_{k=1}^{n} \bar{n}_{1k} = 1$, meaning that the target flow is $1$, as it should be.

Note from Equation (\ref{Eq_forward_backward_variables01}) that, as edge $(1,k)$ ($k\ne 1$) does not appear on any path in $\mathcal{P}_{kn}$ by construction, the backward variables $z'_{kn}$ with $k>1$ do not depend on the $\lambda_{k}^{\mathrm{in}}$. Thus, from (\ref{Eq_capacity_equality_constraints01}), for updating the $\lambda_{k}^{\mathrm{in}}$, we have to solve
\begin{equation}
\frac{ p_{1k}^{\mathrm{ext}} \exp[-\theta \lambda_{k}^{\mathrm{in}}]  z'_{kn} } {\sum_{k' \in \mathcal{I}n} p_{1k'}^{\mathrm{ext}} \exp[-\theta \lambda_{k'}^{\mathrm{in}}]  z'_{k'n}} = \sigma_{k}^{\mathrm{in}} \quad \text{for all } k \in \mathcal{I}n
\label{Eq_equation_with_lambda_in01}
\end{equation}
These equations are invariant when adding a constant to the Lagrange parameters. Therefore, we will impose the natural condition $\sum_{k \in \mathcal{I}n} \lambda_{k}^{\mathrm{in}} \sigma_{k}^{\mathrm{in}} = 0$, with the consequence that we have the following nice property: the expected augmented cost from source node 1 to target node $n$ is equal to the real expected cost ($\langle \tilde{c}' \rangle = \langle \tilde{c} \rangle$).

It was shown in the appendix of \cite{Lebichot-2018} that the solution of this system of logistic function equations is
$\lambda_{k}^{\mathrm{in}} = -\tfrac{1}{\theta} \log \big( \frac{\sigma_{k}^{\mathrm{in}}} { p_{1k}^{\mathrm{ext}}  z'_{kn} } \big) - \sum_{l \in \mathcal{I}n} \sigma_{l}^{\mathrm{in}} \big[ -\tfrac{1}{\theta} \log \big( \frac{\sigma_{l}^{\mathrm{in}}} { p_{1l}^{\mathrm{ext}}  z'_{ln} } \big) \big]$ and because we know that $p_{1l}^{\mathrm{ext}} = \sigma_{l}^{\mathrm{in}}$, we further obtain
\begin{equation}
\lambda_{k}^{\mathrm{in}} = \tfrac{1}{\theta} \bigg( \log z'_{kn} - \sum_{l \in \mathcal{I}n} \sigma_{l}^{\mathrm{in}} \log z'_{ln} \bigg)
= \sum_{l \in \mathcal{I}n} \sigma_{l}^{\mathrm{in}} (\phi_{l} - \phi_{k})  \quad \text{ for } k \in \mathcal{I}n
\end{equation}
where we used Equation (\ref{Eq_optimal_free_energy01}) and $\sum_{l \in \mathcal{I}n} \sigma_{l}^{\mathrm{in}} = 1$. It can be observed that the $\log z'_{kn}$ is centered with respect to the weighted mean (with weights provided by $\sigma_{k}^{\mathrm{in}}$). This last expression proves the first part of Equation (\ref{Eq_lagrange_parameters_updates01}).

\paragraph{Computing Lagrange parameters associated to output nodes.}

Symmetrically to (\ref{Eq_number_passages_in01}), we have for the Lagrange parameters $\{ \lambda_{l}^{\mathrm{out}} \}_{l \in \mathcal{O}ut}$ associated to the edges incident to the target node $n$, 
\begin{equation}
\bar{n}_{ln} = \frac{ z'_{1l} p_{ln}^{\mathrm{ext}} \exp[-\theta c'_{ln}]  z'_{nn} } {z'_{1n}}
= \frac{ z'_{1l} p_{ln}^{\mathrm{ext}} \exp[-\theta \lambda_{l}^{\mathrm{out}}] } {\sum_{l' \in \mathcal{O}ut} z'_{1l'}  p_{l'n}^{\mathrm{ext}} \exp[-\theta \lambda_{l'}^{\mathrm{out}}] }
\label{Eq_number_passages_out01}
\end{equation}
because $z'_{nn} = 1$. As before, $\sum_{l=1}^{n} \bar{n}_{ln} = 1$ is verified. Setting $\bar{n}_{ln} = \sigma_{l}^{\mathrm{out}}$ and proceeding in the same way as before provides, for the Lagrange parameters in $\mathcal{O}ut$,
\begin{equation}
\lambda_{l}^{\mathrm{out}} = \tfrac{1}{\theta} \Bigg( \log z'_{ln} - \log \bigg( \dfrac{\sigma_{l}^{\mathrm{out}}} {  p_{ln}^{\mathrm{ext}} } \bigg) - \sum_{k \in \mathcal{O}ut} \sigma_{k}^{\mathrm{out}} \bigg[ \log z'_{kn} - \log \bigg( \dfrac{\sigma_{k}^{\mathrm{out}}} {  p_{kn}^{\mathrm{ext}} } \bigg) \bigg] \Bigg) \text{ for } l \in \mathcal{O}ut 
\label{Eq_lagrange_parameters_update_destination01}
\end{equation}
We observe that both $\sum_{k \in \mathcal{I}n} \lambda_{k}^{\mathrm{in}} \sigma_{k}^{\mathrm{in}} = 0$ and $\sum_{l \in \mathcal{O}ut} \lambda_{l}^{\mathrm{out}} \sigma_{l}^{\mathrm{out}} = 0$.
This further proves the second part of Equation (\ref{Eq_lagrange_parameters_updates01}).

These Lagrange parameters for input nodes and output nodes are recomputed in turn until convergence. They are then used in order to calculate the final augmented cost values and the corresponding fundamental matrix, allowing to obtain the quantities of interest, in particular, the optimal transition probabilities matrix (the policy).

\bibliographystyle{myabbrv}
\bibliography{./Biblio.bib}

\begin{center}
\rule{2.5in}{0.01in}
\end{center}

\end{document}